%% file: main.tex
\documentclass[journal,onecolumn,12pt]{IEEEtran}
\usepackage{graphicx}

\usepackage{wrapfig}
\usepackage{tikz}
\usepackage{comment}
\usepackage{amsmath,amssymb} 
\usepackage{color}

\usepackage[left=1in,right=1in,top=1in,bottom=1in]{geometry}
\usepackage{array}

\usepackage{cite}

\usepackage{times}

\usepackage{soul}
\usepackage{url}
\usepackage[utf8]{inputenc}
\usepackage{booktabs}
\usepackage{lipsum}
\usepackage{multirow}
\usepackage{verbatim} 
\DeclareMathOperator*{\argmax}{arg\,max}

\usepackage{algorithm}
\usepackage{algorithmic}
\usepackage{subfigure}

\begin{document}

\title{Federated Momentum Contrastive Clustering}



\author{Runxuan Miao and Erdem Koyuncu \\
 \small\smallskip Department of Electrical and Computer Engineering \\ \small University of Illinois at Chicago\\ \small E-mail: \{rmiao6, ekoyuncu\}@uic.edu}


%


\maketitle

\input{00-abs}


%

\input{01-intro}

\input{02-related}
\input{03-model}

\input{04-exp}

\input{05-ablation}

\input{035-memefficientsupp}

\input{06-conc}

\section*{Acknowledgement}
This work was supported in part by Army Research Lab (ARL) under Grant W911NF-21-2-0272, National Science Foundation (NSF) under Grant CNS-2148182, and by an award from the University of Illinois at Chicago Discovery Partners Institute Seed Funding Program.




%


\bibliographystyle{IEEEtran}
\bibliography{cite}
\end{document}

%% file: 00-abs.tex
\begin{abstract}\vspace{-5pt}
We present federated momentum contrastive clustering (FedMCC), a learning framework that can not only extract discriminative representations over distributed local data but also perform data clustering. In FedMCC, a transformed data pair passes through both the online and target networks, resulting in four representations over which the losses are determined. The resulting high-quality representations generated by FedMCC can outperform several existing self-supervised  learning methods for linear evaluation and semi-supervised learning tasks. FedMCC can easily be adapted to ordinary centralized clustering through what we call momentum contrastive clustering (MCC). We show that MCC achieves state-of-the-art clustering accuracy results in certain datasets such as STL-10 and ImageNet-10. We also present a method to reduce the memory footprint of our clustering schemes.\vspace{-5pt}
\end{abstract}


%% file: 01-intro.tex
\section{Introduction}
\subsection{Federated Learning}
The number of resource-limited mobile and Internet of Things (IoT) devices have increased exponentially over the last decade \cite{alsharif2017green}. With increasing demand for data privacy and prevalence of edge computing devices, the emerging federated learning methods enable training machine learning models over many computationally and power restricted nodes \cite{fedavg-17-mcmahan, li2020federated, kairouz2021advances}. In a federated learning framework, a central server typically keeps track of a global neural network model, which is updated via the local clients through averaging. The local clients update the global model via their own local data.

Numerous works have been proposed to resolve the challenges of federated learning including communication/computation costs \cite{jakub-17, luo2021cost, fedprox-20-li}, device limitations \cite{li2021model}, non-identical distribution of local data \cite{fed-18-zhao, moon-21-cvpr}, privacy concerns \cite{truex2019hybrid}, and node resilience \cite{koyuncu:c26}.
In particular, \cite{jakub-17} optimizes the communication efficiency by reducing the rate of communicated variables using methods like quantization. Early exit methods \cite{kaya2019shallow, koyuncu:c31} have also been utilized to reduce the computation costs in federated learning \cite{zhong2022flee}. MOON~\cite{moon-21-cvpr} addresses the unbalanced data distribution problem via model-level contrastive learning to maximize the similarity between global and local models. 
However, these supervised federated learning methods require large amounts of manually labelled data and are typically unable to cluster samples.

\subsection{Federated Representation Learning}
Recently, effort has focused on unsupervised representation learning methods such as SimCLR~\cite{simclr-20-chen}, BYOL~\cite{byol-20}, MoCo~\cite{moco-20-he}, and SimSiam~\cite{simsiam-21-chen}. 
Contrastive learning has been a key technique in representation learning, but it requires a large amount of negative pairs.
SimCLR and MoCo generate negative pairs from a large batch size and a memory bank, respectively. 
For example, NCC~\cite{NCC-21-huang} maps clusters to an embedding space,
and MiCE\cite{mice-21-wei} uses the instance discrimination to learn semantic clusters. 
However, these methods can only be applied if the data is centralized to one device.

Some existing methods explore federated training with unlabeled data, but they are primarily designed for linear evaluation or semi-supervised learning, hence not productive in clustering tasks. 
For instance, \cite{FedCA-20-zhang, fedu-21-zhuang} consider federated unsupervised representation learning to address the inconsistency of representation space across clients.
FedCA~\cite{FedCA-20-zhang} introduces a dictionary with an alignment module, while FedU\cite{fedu-21-zhuang} and FedEMA \cite{fedema-2022} propose a divergence-aware predictor update. 


\subsection{Federated Clustering}
A key application area of unsupervised  learning is clustering, which aims at grouping data points without labels. There are various proposed schemes for the classical centralized clustering task \cite{TCC-21-shen, NCC-21-huang,cc-21-li, mice-21-wei, idfd-21-tao, embedding-16-xie, Yang_2016_CVPR, Chang_2017_ICCV, Wu_2019_ICCV, Ji_2019_ICCV}, most of which are based on contrastive learning. Beside these works on centralized clustering, we also explore recent works focusing on federated clustering~\cite{uc-Berkekey-20, fl-hc-20-fan, patient-19-huang, matrix-factorization-20-wang}.
For example, an iterative federated clustering algorithm~\cite{uc-Berkekey-20} focuses on distributing and partitioning clients into clusters but do not focus on clustering data. The work \cite{fl-hc-20-fan} proposes hierarchical clustering of local updates to improve convergence of federated learning. Additionally, clustering distributed data collected from private medical records~\cite{patient-19-huang} helps medical treatment in hospitals. Some classical $k$-means based or variant methods will have difficulties with complex data features and/or large number of clusters~\cite{matrix-factorization-20-wang, clustering-02-dhillon, scalable-12-bahman, para-kmeans-14-kucu}. For example, federated matrix factorization~\cite{matrix-factorization-20-wang} has been introduced for data clustering.
However, their experiments contain only synthetic data or simple dataset such as MNIST~\cite{mnist}.



\subsection{Summary of Contributions}

In this paper, we propose federated momentum contrastive clustering (FedMCC) and centralized momentum contrastive clustering (MCC) based on CC~\cite{cc-21-li} and BYOL~\cite{byol-20}.  
Our schemes provide high-quality representations amenable to clustering in both federated and centralized settings.
Similar to BYOL~\cite{byol-20} and CC~\cite{cc-21-li}, we conduct instance- and cluster- level contrastive learning between two neural networks: a trained online network and a target network with a slow-moving average of the online parameters. In FedMCC, a transformed data pair passes through both the online and target networks, resulting in four representations over which the losses are determined. The corresponding loss function is akin to the symmetrized loss functions in \cite{caron2020unsupervised, chen2021empirical}. This way, our method encourages more information to be encoded to online and target networks. FedMCC does not rely on large batch sizes~\cite{simclr-20-chen} or memory banks~\cite{moco-20-he}. 

We summarize the main contributions of our work as follows: 



\begin{itemize}
\item We make the first attempt towards studying contrastive representation learning for federated clustering by proposing federated momentum contrastive clustering (FedMCC) scheme, which is based on the BYOL~\cite{byol-20} and CC~\cite{cc-21-li}. The proposed FedMCC outperforms various baselines by comfortable margins.
\item Our scheme not only addresses the difficult problem of training a clustering scheme in a distributed manner but also achieves state-of-the-art results on learning representation for linear evaluation or semi-supervised settings. This is in contrast to many existing federated learning schemes that are designed for a specific  task.
\item FedMCC can even be adapted to ordinary centralized clustering. The resulting MCC scheme outperforms all existing methods on STL-10 and ImageNet-10 datasets. 
\item We also describe an algorithm to reduce the memory footprint of our schemes.
\end{itemize}
Overall, our proposed FedMCC scheme is a simple and effective federated clustering framework for edge devices that can be tailored for various other learning tasks. 

\subsection{Organization}
The rest of the paper is organized as follows: In Sections \ref{secCentClust} and \ref{secFedClust}, we introduce the MCC and FedMCC schemes, respectively. Numerical experiments are provided in Section \ref{sec:exp}, and ablation studies are reported in Section \ref{secablation}. We discuss how to design memory-efficient schemes in Section \ref{secmemefficient}. Finally, in Section \ref{secconclusions}, we draw our main conclusions.

%% file: 03-model.tex
\section{Centralized Clustering}
\label{secCentClust}
In this section, we introduce our centralized clustering scheme. We will later extend our centralized clustering scheme to federated learning in Section \ref{secFedClust}. Our centralized clustering scheme, which we will refer to as Momentum Contrastive Clustering (MCC), relies on a momentum extension of the contrastive clustering (CC) scheme that was proposed in \cite{cc-21-li}. We thus provide an overview of CC to properly motive MCC. 


\subsection{Contrastive Clustering (CC)}
\begin{figure}
\centering
\includegraphics[width=.6\columnwidth]{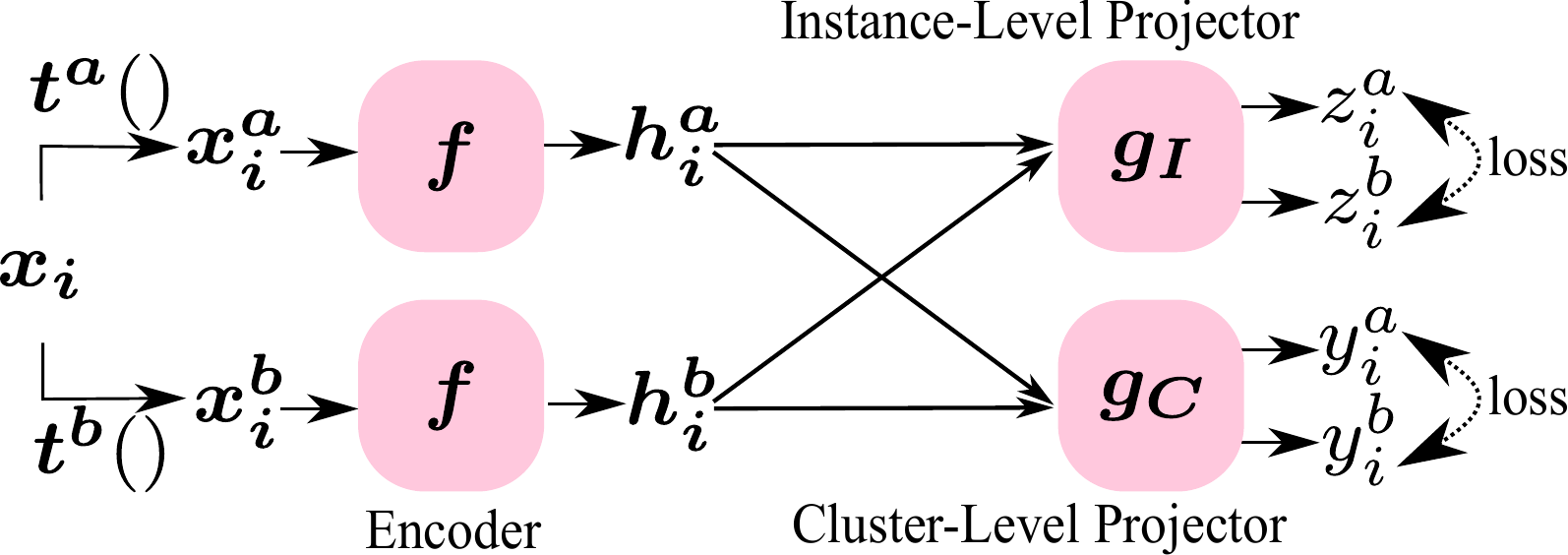}
\caption{The CC model}
\label{fig:cc}
\end{figure}

The block diagram of the CC scheme is shown in Fig. \ref{fig:cc}. We consider a dataset $\mathcal{D} = \{x_1,\ldots,x_{|\mathcal{D}|}\}$. Given an input $x_i\in\mathcal{D}$, the CC scheme first creates two samples $x_i^{a} \triangleq t^{a}(x_i)$ and $x_i^{b} \triangleq t^{b}(x_i)$ through transformations $t^{a}$ and $t^{b}$, respectively. We use the variable $\sigma\in\{a,b\}$ to represent the sample index so that the transformations are succinctly expressed as  $x_i^{\sigma} \triangleq t^{\sigma}(x_i),\,\sigma\in\{a,b\}$.
The transformations are sampled uniformly at random from a family $\mathcal{T}$ of augmentations, which may include rotations, noise, etc. The samples then pass through the same encoder $f$, creating feature vectors $h_i^{\sigma} \triangleq f(x_i^{\sigma}),\,\sigma\in\{a,b\}$. 
An instance-level multi-layer perceptron (MLP) $g_I$ projects $h_i^a$ and $h_i^b$ to obtain instance-level representations $z_i^{\sigma} \triangleq g_I(h_i^{\sigma})\in \mathbb{R}^{d_1},\,\sigma\in\{a,b\}$. Likewise, a cluster-level MLP $g_C$ produces cluster-level representations $y_i^{\sigma} \triangleq g_C(h_i^{\sigma})\in \mathbb{R}^{d_2},\,\sigma\in\{a,b\}$. In the CC scheme, the output dimensionality $d_2$ of the cluster-level representations is chosen to be equal to the number of clusters one wishes to find in the dataset. In many cases, the instance-level output dimensionality $d_1$ is chosen to be much larger than $d_2$.

The similarity of any two representations are compared via the cosine similarity measure $ s(u, v) \triangleq u^{\dagger} v /(\|u\|\|v\|)$. To define the loss functions, we need the following definitions. Given matrices $\mathbf{u} = [u_1 \cdots u_N]\in\mathbb{R}^{d\times n}$ and $\mathbf{v} = [v_1 \cdots v_N]\in\mathbb{R}^{d\times n}$ constructed via the indicated column vectors, we define the contrastive loss function
\begin{align}
\label{mainloss}
L(\mathbf{u},\mathbf{v};\tau) \triangleq \frac{1}{n} \sum_{i=1}^n
   -\log\frac{\exp\bigl(\frac{1}{\tau}s(u_i, v_i)\bigr)}{\sum_{\substack{j=1 \\ j \neq i}}^n \left[\exp\bigl(\frac{1}{\tau}s(u_i,u_j)\bigr) \!+\! \exp\bigl(\frac{1}{\tau}s(u_i, v_j)\bigr) \right]}.\!\!
\end{align}
Given a batch size $n$, the instance-level contrastive loss is then defined via the instance-level representations $\mathbf{z}^{\sigma} \triangleq [z_1^{\sigma} \cdots z_n^{\sigma}]\in\mathbb{R}^{d_1\times n},\,\sigma\in\{a,b\}$ as  $
    L(\mathbf{z}^a, \mathbf{z}^b; \tau_I)$,
where $\tau_I > 0$ is the instance-level temperature parameter. On the other hand, given $\mathbf{c}^{\sigma} \triangleq [y_1^{\sigma} \cdots y_n^{\sigma}]^{\dagger}\in\mathbb{R}^{n\times d_2},\,{\sigma}\in\{a,b\}$, the cluster-level contrastive loss is defined by $
    L(\mathbf{c}^a, \mathbf{c}^b; \tau_C)$,
 where $\tau_C > 0$ is the cluster-level temperature.
The cluster-level contrastive loss is utilized to differentiate clusters. Specifically, columns of
$\mathbf{c}^a$ and $\mathbf{c}^b$ are considered as the representation of each cluster. On the other hand, rows of
$\mathbf{c}^a$ and $\mathbf{c}^b$ (i.e. $y_i^{\sigma}$s) correspond to the soft labels of samples. In particular, in a deterministic assignment of inputs to clusters, all rows would be one-hot encoded vectors.



The overall loss function of CC is a entropy-regularized linear combination of instance-level and cluster-level losses. In precise form, the loss function is given by
\begin{align}
\label{lccloss}
    L_{CC} \!\triangleq\! \tfrac{1}{2}(L(\mathbf{z}^a, \mathbf{z}^b; \tau_I) \!+\!  L(\mathbf{c}^a,\mathbf{c}^b; \tau_C))\!+\! H(\mathbf{c}^a)\!+\!H(\mathbf{c}^b), \!\!
\end{align}
where, for any matrix $\mathbf{u} = [u_1 \cdots u_d]\in\mathbb{R}^{n\times d}$, the entropy is defined as
\begin{align}
    H(u) \triangleq -\sum_{i=1}^{d} \frac{\|u_i\|_1}{\|\mathbf{u}\|_1}\log \frac{\|u_i\|_1}{\|\mathbf{u}\|_1}.
\end{align}
As discussed in \cite{cc-21-li}, entropy regularization helps avoid the trivial solution where all samples are assigned to the same cluster.



\subsection{Momentum Contrastive Clustering (MCC)}
\label{mccscheme}
\begin{figure}[ht]
\centering
\includegraphics[width=.7\columnwidth]{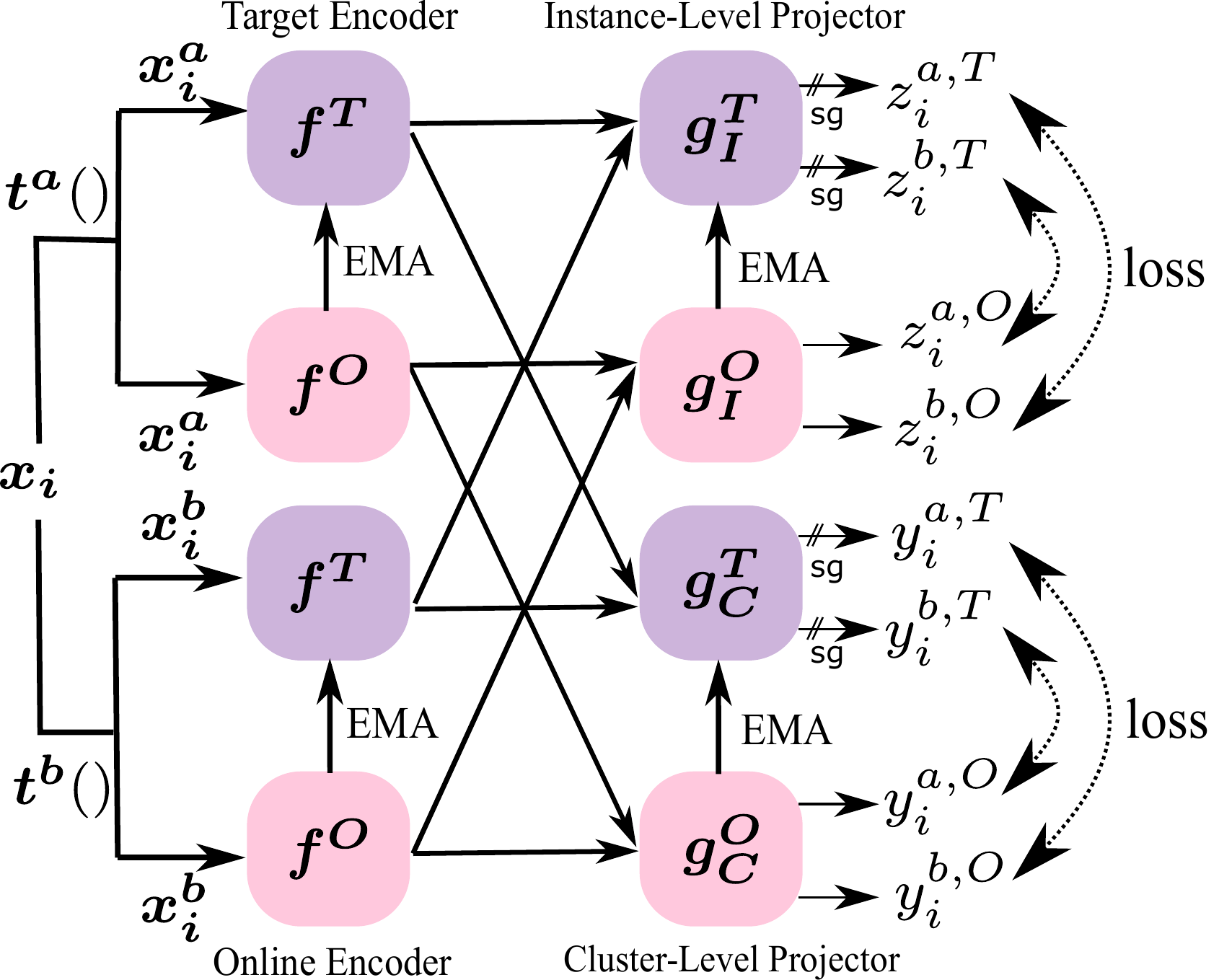}
\caption{The MCC architecture
}
\label{fig:mcc}
\end{figure}
We introduce MCC as an extension of CC with momentum. Numerical results in Section \ref{sec:exp} will show that the resulting MCC scheme not only achieves the best clustering accuracy performance available in the literature for certain datasets, but also improves the performance of CC for any given dataset. 


The block diagram of our proposed MCC scheme is shown in Fig. \ref{fig:mcc}. Our idea is to replicate each block in the CC scheme to two different ``online'' and ``target'' versions. This idea was originally proposed in the context of contrastive learning via BYOL \cite{byol-20}, but  \cite{byol-20} utilizes the mean-squared error (MSE) loss function. MSE loss provides poor performance in our settings. We shall thus utilize instead the loss in (\ref{mainloss}). Also, in the original BYOL scheme \cite{byol-20}, and the follow-up variants such as \cite{fedu-21-zhuang, fedema-2022}, one sample is fed to the online network while the other is fed to the target network. A key novelty of our architecture is that both samples are fed to both networks. This diversity results in a symmetric loss function and improves the overall performance. 


In detail, given input $x_i \in \mathcal{D}$, we create the two samples $t^{\sigma}(x_i),\,\sigma\in\{a,b\}$, similar to the the CC scheme. The two samples are processed by what we call an ``online network'' and a ``target network'' simultaneously. We use the superscripts ``$O$'' and ``$T$'' to represent variables related to online and target networks, respectively. 
The online network consists of an encoder $f^O$, an instance-level projector $g_I^O$, and a cluster-level projector $g_C^O$. 
The target network has the same architecture as the online network; consisting of an encoder $f^T$, an instance-level projector $g_I^T$, and a cluster-level projector $g_C^T$. The building blocks of the online and the target networks are represented by boxes with different colors in Fig. \ref{fig:mcc}.  The parameters of the target network are calculated by using an exponential moving average (EMA) of the  parameters of the online network.


Given $\sigma\in\{a,b\}$ representing the sample index as before, and $\eta\in\{O,T\}$ representing the network index, the instance-level and cluster-level representations generated from the online and target networks are defined  as
\begin{align}
\label{mccdef1}
 z_{i}^{\sigma,\nu} \triangleq  g_I^{\nu}(f^{\nu}(t^{\sigma}(x_{i}))),\,  
y_{i}^{\sigma,\nu} \triangleq  g_C^{\nu}(f^{\nu}(t^{\sigma}(x_{i}))),\,\scalebox{0.9}{$\sigma\!\in\!\{a,b\},\,\nu\!\in\!\{O,T\}$}.
\end{align}
To define the loss function, we collect the representation vectors into matrices
\begin{align}
\label{mccdef2}
 \mathbf{z}^{\sigma, \nu} \!\triangleq\! [z_1^{\sigma, \nu} \!\cdots z_n^{\sigma, \nu}]\!\in\!\mathbb{R}^{d_1\times n}\!,\,
        \mathbf{c}^{\sigma, \nu} \!\triangleq\! [y_1^{\sigma, \nu} \!\cdots y_n^{\sigma, \nu}]^{\dagger}\!\in\!\mathbb{R}^{n\times d_2}\!,\,\scalebox{0.9}{$\sigma\!\in\!\{a,b\},\,\nu\!\in\!\{O,T\}$}.\!\!
\end{align}

Losses are evaluated across the outputs of the online and target networks for different augmentations and indicated by arrows in Fig. \ref{fig:mcc}. As in the CC scheme, we apply entropy regularization to the cluster-level representations. Mathematically, for a given batch, the loss function for MCC is expressed as
\begin{multline}
\label{mcclossfunc}
    L_{MCC} \triangleq \tfrac{1}{2}(L(\mathbf{z}^{a, O}, \mathbf{z}^{b, T}; \tau_I) + L(\mathbf{z}^{a, T}, \mathbf{z}^{b, O}; \tau_I)  +
     L(\mathbf{c}^{a, O},\mathbf{c}^{b, T}; \tau_C) + \\ L(\mathbf{c}^{a, T},\mathbf{c}^{b, O}; \tau_C))+ H(\mathbf{c}^{a, O})+
     H(\mathbf{c}^{b, O}) + H(\mathbf{c}^{a, T})+H(\mathbf{c}^{b, T}).
\end{multline}

At each epoch and each batch update, only the parameters of the online networks are updated via gradient descent, and the target network parameters are kept frozen. The target network parameters are then updated via an EMA filter via $\mathcal{P}(\nu^T) \leftarrow 
m\mathcal{P}(\nu^T) + (1-m)\mathcal{P}(\nu^O),\,\nu\in\{f,g_C,g_I\}$, where $\mathcal{P}(\cdot)$ represents the parameter set of its argument. Also, the momentum factor or the decay rate $m\in(0,1)$ is a hyperparameter to be set and fixed before one initiates gradient descent updates.



\section{Federated Clustering}
\label{secFedClust}
\begin{figure}[ht]
\centering
\includegraphics[width=.9\columnwidth]{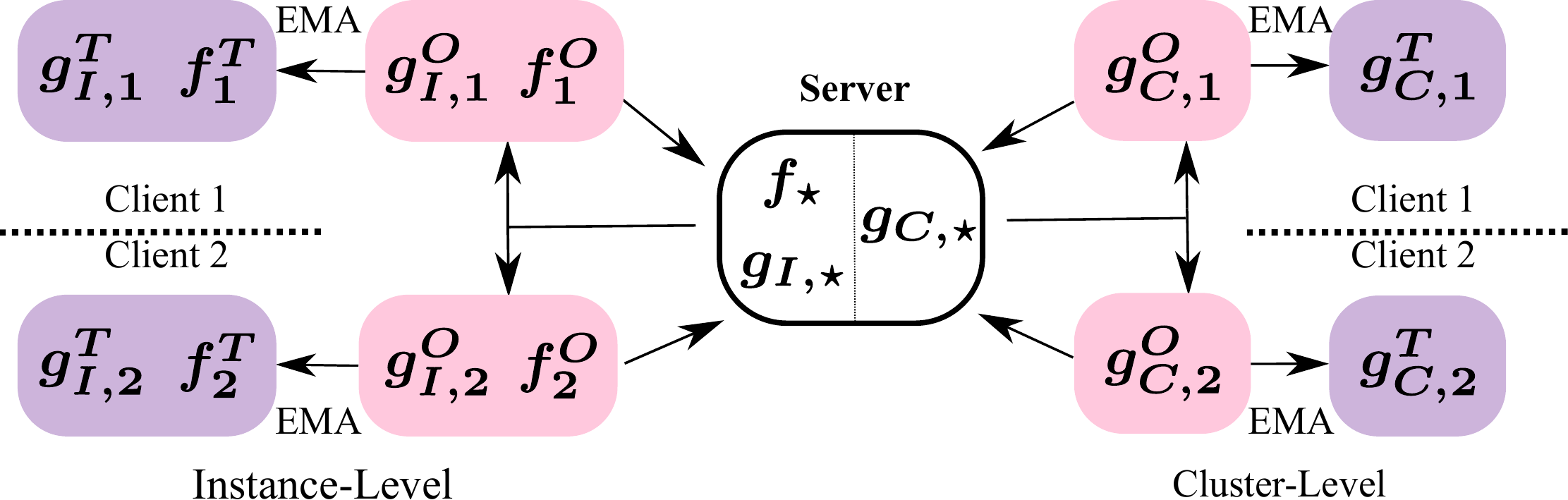}
\caption{The FedMCC scheme for the special case of two clients}
\label{fig:federated}
\end{figure}
\subsection{Problem Statement and Formulation}
We now consider the case of federated clustering.
Suppose there are $K$ clients, where Client $k$ has its local unlabeled data $\mathcal{D}_k$. 
Our goal is to learn a machine model over the dataset $\mathcal{D} \triangleq \bigcup_{k=1}^K \mathcal{D}_k$ on the central server. Due to the demand for data privacy, clients can not send their local data to a central sever to train a model. 
We aim to learn a global clustering model by training and aggregating models trained at each client with local data. Our idea is to extend the MCC scheme in Section \ref{mccscheme} to the federated setting. We refer to the resulting scheme by Federated MCC (FedMCC).


Let $\mathcal{D}_k \triangleq \{x_{1,k},\ldots,x_{|\mathcal{D}_k|,k}\},\,k=1,\ldots,K$ represent the local datasets of the users. Each user or client implements the MCC architecture in Fig. \ref{fig:mcc}. Let $t_k^{\sigma},f_k^{\nu},g_{C,k}^{\nu}$, $g_{I,k}^{\nu},\,\sigma\in\{a,b\},\,\nu\in\{O,T\}$ denote the MCC network elements at user $k$. The central server also has the same MCC structure, containing the ``global'' networks $f_{\star},g_{I,\star}$, and $g_{C,\star}$. The overall block diagram of the FedMCC architecture is shown in Fig. \ref{fig:federated}. 

To define the loss functions, we simply extend the notation in (\ref{mccdef1}) and (\ref{mccdef2}) to multiple users by defining, for each user $k\in\{1,\ldots,K\}$, $\sigma\in\{a,b\}$, and $\nu\in\{O,T\}$, the representation vectors 
\begin{align}
z_{i,k}^{\sigma,\nu} \triangleq  g_{I,k}^{\nu}(f_k^{\nu}(t_k^{\sigma}(x_{i,k}))),\,
 y_{i,k}^{\sigma,\nu} \triangleq  g_{C,k}^{\nu}(f_k^{\nu}(t_k^{\sigma}(x_{i,k}))),
\end{align}
and their matricized versions $
 \mathbf{z}_k^{\sigma, \nu} \triangleq [z_{1,k}^{\sigma, \nu} \!\cdots z_{n,k}^{\sigma, \nu}]$, and $ \mathbf{c}_k^{\sigma, \nu} \!\triangleq\! [y_{1,k}^{\sigma, \nu} \!\cdots y_{n,k}^{\sigma, \nu}]^{\dagger}$. 

We now define the instance-level contrastive loss of user $k$ as
\begin{align}
\label{eq:fed_Lins}
    & L_{I,k} \triangleq \tfrac{1}{2}(L(\mathbf{z}_k^{a, O}, \mathbf{z}_k^{b, T}; \tau_I) + L(\mathbf{z}_k^{a, T}, \mathbf{z}_k^{b, O}; \tau_I)),
\end{align}
and the cluster-level contrastive loss of user $k$ as
\begin{align}
\label{eq:fed_Lclu}
    L_{C,k} \triangleq & \tfrac{1}{2}(L(\mathbf{c}_k^{a, O},\mathbf{c}_k^{b, T}; \tau_C) + L(\mathbf{c}_k^{a, T},\mathbf{c}_k^{b, O}; \tau_C))+ H(\mathbf{c}_k^{a, O})+ 
     H(\mathbf{c}_k^{b, O}) + H(\mathbf{c}_k^{a, T})+H(\mathbf{c}_k^{b, T}).
\end{align}



\subsection{Pipeline of FedMCC}
Note that the sum of the two losses $L_{I,k}+ L_{C,k}$ in (\ref{eq:fed_Lins}) and (\ref{eq:fed_Lclu}) takes the same form as the loss function for MCC in (\ref{mcclossfunc}). A first idea for designing a federated MCC scheme is therefore to optimize the loss $L_{I,k}+L_{C,k}$ at each user. However, it turns out that this strategy results in poor performance. In fact, cluster-level representations typically have much lower dimensionality than instance-level representations. We thus expect that training consistent cluster-level representations across multiple users with heterogeneous datasets is much more difficult than training consistent instance-level representations. Hence, a key idea of this paper is to advocate a multi-stage solution to the federated clustering problem. One should first construct consistent high-dimensional representations across users, which can ultimately be reduced to cluster representations in a consistent manner. In the case of an extension of MCC, this is done by breaking apart the centralized MCC cost function into the two parts shown in (\ref{eq:fed_Lins}) and (\ref{eq:fed_Lclu}). 

Corresponding to the two loss functions in (\ref{eq:fed_Lins}) and (\ref{eq:fed_Lclu}), FedMCC relies on a two-stage learning scheme. In the first representation learning stage, the clients optimize the loss function (\ref{eq:fed_Lins}), and the learned models are combined at the server every $E$ epochs through federated averaging. Once the first stage converges, in the second clustering stage, the clients optimize the loss function in (\ref{eq:fed_Lclu}). A detailed algorithmic description of the two stages is provided in Algorithms \ref{alg:FedMCC_1st} and \ref{alg:FedMCC_2nd}, respectively. Algorithm \ref{alg:FedMCC_1st} aims to learn the global encoder $f_{\star}$ and the global instance-level projector $g_{I,\star}$, and the cluster-level projectors are irrelevant to the loss functions. Algorithm \ref{alg:FedMCC_2nd}, on the other hand, learns the global cluster-level projector $g_{C,\star}$ given the global encoder $f_{\star}$ in the first stage. The global encoder $f_{\star}$ remains frozen throughout Algorithm \ref{alg:FedMCC_2nd}. In the test stage, only $f_{\star}$ and $g_{C,\star}$ are needed to determine the cluster index of a given input.

\begin{algorithm}
	\caption{FedMCC: Representation Learning}
	\label{alg:FedMCC_1st}
	\begin{algorithmic}[1]
		\renewcommand{\algorithmicrequire}{\textbf{Input:}}
		\renewcommand{\algorithmicensure}{\textbf{Output:}}

		\REQUIRE Number of communication rounds $R$, Number of clients $K$, Number of local epochs $E$.
		\ENSURE  Global encoder $f_{\star}$
		
		\STATE{\textbf{Server executes:} Initialize server's network parameters $f_{\star}$ and $g_{I,\star}$.}
		\FOR{$r = 1, \ldots, R$}
		    \FOR{$k = 1, 2, \ldots, K$ in parallel}
		    \STATE{Send global encoder $f_{\star}$ and global instance-level projector $g_{I,\star}$ to client $k$.}
		    \STATE{$f_{k}$, $g_{I,k}$ $\leftarrow$ \textbf{ClientLocalTraining}($k$, $f_{\star}$, $g_{I,\star}$)}
		    \ENDFOR
		   \STATE{Federated averaging: $\mathcal{P}(f_{\star})\leftarrow\sum_{k=1}^K \frac{|\mathcal{D}_k|}{|\mathcal{D}|}\mathcal{P}(f_{k}),\,\mathcal{P}(g_{I,\star})\leftarrow\sum_{k=1}^K \frac{|\mathcal{D}_k|}{|\mathcal{D}|}\mathcal{P}(g_{I,k})$}.
		\ENDFOR
		\STATE{Return global encoder $f_{\star}$.}
		\STATE{\textbf{ClientLocalTraining}($k$,  $f_k$, $g_{I, k}$) }
		\FOR{$\mbox{epochs} = 1,\ldots,E$ and size-$n$ batch learning within each epoch over dataset $\mathcal{D}_k$}
		\STATE{Update the online networks $f_k^O$ and $g_{I, k}^O$ of client $k$ by descending the gradient of the instance-level contrastive loss $L_{I,k}$ in \eqref{eq:fed_Lins}.}
		\STATE{Update the target network parameters $f_k^T$ and $g_{I, k}^T$ of client $k$ via EMA.}
		\ENDFOR
		\STATE{Return the online networks $f_k^O$ and $g_{I, k}^O$.}
	\end{algorithmic}
\end{algorithm}

\begin{algorithm}
	\caption{FedMCC: Clustering}
	\label{alg:FedMCC_2nd}
	\begin{algorithmic}[1]
		\renewcommand{\algorithmicrequire}{\textbf{Input:}}
		\renewcommand{\algorithmicensure}{\textbf{Output:}}

		\REQUIRE Number of communication rounds $R$, Number of clients $K$, Number of local epochs $E$. A pre-trained (via Algorithm \ref{alg:FedMCC_1st}) global encoder $f_{\star}$ available at each client $f_k = f_{\star},\,\forall k$.
		\ENSURE  Global cluster projector $g_{C,\star}$, cluster assignments.
		
		\STATE{\textbf{Server executes:} Initialize $g_{C,\star}$.}
		\FOR{$r = 1, \ldots , R $}
		    \FOR{$k = 1, 2, \ldots, K$ in parallel}
		    \STATE{Send global cluster-level projector $g_{C,\star}$ to client $k$.}
		    \STATE{$g_{C,k}$ $\leftarrow$ \textbf{ClientLocalTraining}($k$, $g_{C,\star}$)}
		    \ENDFOR
		   \STATE{Federated averaging: $\mathcal{P}(g_{C,\star})\leftarrow\sum_{k=1}^K \frac{|\mathcal{D}_k|}{|\mathcal{D}|}\mathcal{P}(g_{C,k})$}
		\ENDFOR
		\STATE{Return global cluster projector $g_{C,\star}$.}
		\STATE{\textbf{Test phase:} Compute the cluster assignment of a test image $x$ by $c \leftarrow \argmax g_{C,\star}(f_{\star}(x))$.\!\!\!}
		\STATE{\textbf{ClientLocalTraining}($k$, $g_{C, k}$)}
		\FOR{$\mbox{epochs} = 1,\ldots,E$ and size-$n$ batch learning within each epoch over dataset $\mathcal{D}_k$}
		\STATE Update the online network $g_{C, k}^O$ of client $k$ by descending the gradient of the cluster-level contrastive loss $L_{C,k}$ in \eqref{eq:fed_Lclu}.
		\STATE{Update the target network $g_{C, k}^T$ of client $k$ via EMA.}
		
		\ENDFOR
		\STATE{Return the online cluster-level projector $g_{C, k}^O$ of client $k$.}

	\end{algorithmic}

\end{algorithm}

%% file: 04-exp.tex
\section{Experiments}
\label{sec:exp}
In this section, we present numerical results that demonstrate the performance of our centralized and federated clustering schemes over different datasets and scenarios. We first present the general experiment setup. We then consider the performance of FedMCC in the federated setting. We then evaluate the quality of representations generated by FedMCC in the linear evaluation and semi-supervised learning settings. Finally, we present centralized clustering results.

\subsection{Experiment Setup}

\textbf{Datasets and Settings:} We have conducted experiments on the CIFAR-10, CIFAR-100, MNIST, STL-10, and Imagenet-10 datasets. The CIFAR-100, CIFAR-100, and MNIST datasets have 10, 100, and 10 classes, respectively. All datasets consist of 50,000 training samples and 10,000 test samples with the same number of data samples per class. We have followed the same train-test split settings for STL-10 and ImageNet10, as in the recent clustering works \cite{cc-21-li,NCC-21-huang,TCC-21-shen}. 
Both datasets contain 13,000 samples and 10 classes. 
We show the performance of our schemes in terms of clustering accuracy (ACC), normalized mutual information (NMI), and adjusted rand index (ARI). Also, following the CC scheme \cite{cc-21-li}, 100,000 unlabeled samples are trained in addition for instance-level representation learning for STL-10.

In the federated settings, for fair comparison with existing methods~\cite{fedu-21-zhuang, FedCA-20-zhang, fedema-2022}, we simulate a centralized node as the server and $K$ distributed nodes as clients.
In the IID scenario, each client contains the same number of images from all classes. In the non-IID scenario, each client has all the samples of $100/K$ classes and no other sample from the rest of the classes. For example, for the CIFAR-10 dataset with $K=2$, clients $1$ and $2$ have all the samples from classes $0$ to $4$ and $5$ to $9$, respectively. 



\textbf{Implementation Details:} For fair comparisons, we use ResNet-34 \cite{resnet-16-he} as the backbone to report the centralized clustering results. 
In the federated setting, following the existing work \cite{fedu-21-zhuang, FedCA-20-zhang, fedema-2022}, we use ResNet-18 or ResNet-50~\cite{resnet-16-he} as encoders and train the model for 100 communication rounds for $K=5$ clients. 
For each communication round, each client is trained for $E=5$ local epochs. 
Additional 10 communication rounds for cluster-level projector (Algorithm~\ref{alg:FedMCC_2nd}) is used for clustering, which is not considered for comparing the performance on linear evaluation and semi-supervised learning.
The Adam optimizer \cite{kingma2014adam} with an initial learning rate of $0.0003$ and no weight decay is used. 
The input size of images is resized to $224 \times 224$ except for the $32 \times 32$ images used for MNIST in both federated and centralized scenarios. 
The output dimension of the instance-level projector is set to 128, and the feature dimension of the cluster-level projector is equal to the number of clusters. 
Unless specified otherwise, we set the batch size to be $n=128$, the instance-level temperature as $\tau_I = 0.5$, the cluster-level temperature as $\tau_C = 1.0$, and the EMA parameter to be $m = 0.99$.



\subsection{Federated Clustering Evaluation}
\label{fedclusteval}
{\bf Baselines:} We consider clustering the representations generated by existing federated learning schemes and compare the resulting performance with that of FedMCC. In particular, we have implemented FedU~\cite{fedu-21-zhuang}, and clustered the resulting representations. We have considered FedU specifically as it does not require extensive hyperparameter tuning and performs very close to state of the art in linear evaluation and semi-supervised learning tasks, as we shall also show later. We have considered ``FedU + K-means,'' which refers to the combination of FedU with the classical  K-means~\cite{k-means-1967} algorithm.  
We also considered the baseline ``FedU + MCC,'' which refers to the combination of FedU with our second stage of federated clustering. Both FedU + CC and FedU + MCC will also be relevant in ablation studies, to be discussed later in Section \ref{secablation}. We update FedU for 100 rounds and update 10 rounds of a cluster-level projector as we do in FedMCC. The baseline ``Fed + CC'' simply trains the CC model~\cite{cc-21-li} via federated averaging. As an upper bound, we have considered the centralized clustering scheme NCC \cite{NCC-21-huang}.



%

{\bf Results: } Tables \ref{tab:federated_iid} and \ref{tab:federated_noniid} show the clustering accuracies and metrics of our FedMCC compared with baselines. We observe that FedMCC consistently outperforms FedU + K-means. Specifically, on the IID CIFAR-10 dataset, there is 6.9\% improvement over FedU-based schemes. We also improve 3.6\% and 2.9\% clustering accuracy on MNIST IID and Non-IID settings, respectively. We also observe that the naive federated generalization of the CC scheme, Fed + CC, is suboptimal in general. The big gap between the performance of FedMCC and the upper bounds is notable, suggesting that there is potentially much room for improvement for the general task of federated clustering. 

\begin{figure*}[t]
\begin{center}
  \subfigure[Round 0]{\label{fig:online-target}\includegraphics[width=30mm]{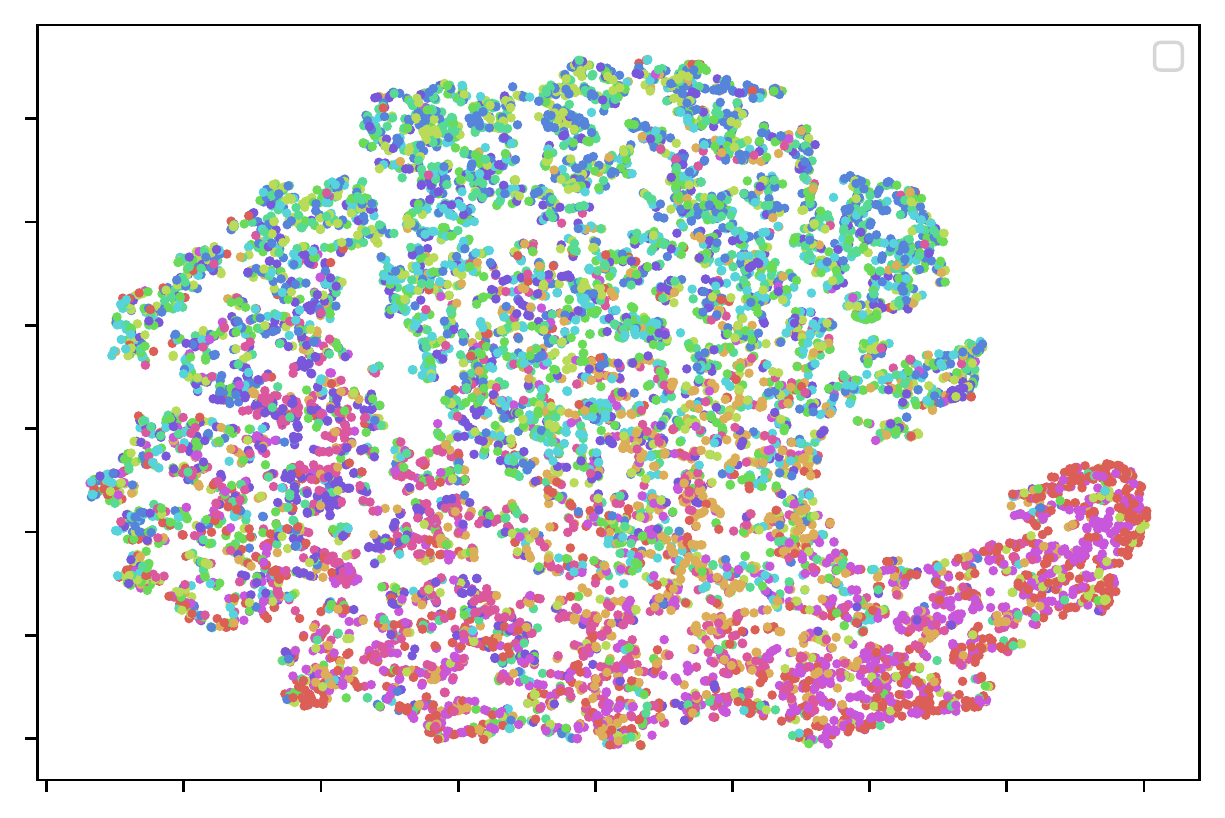}}
  \subfigure[Round 10]{\label{fig:target-online}\includegraphics[width=30mm]{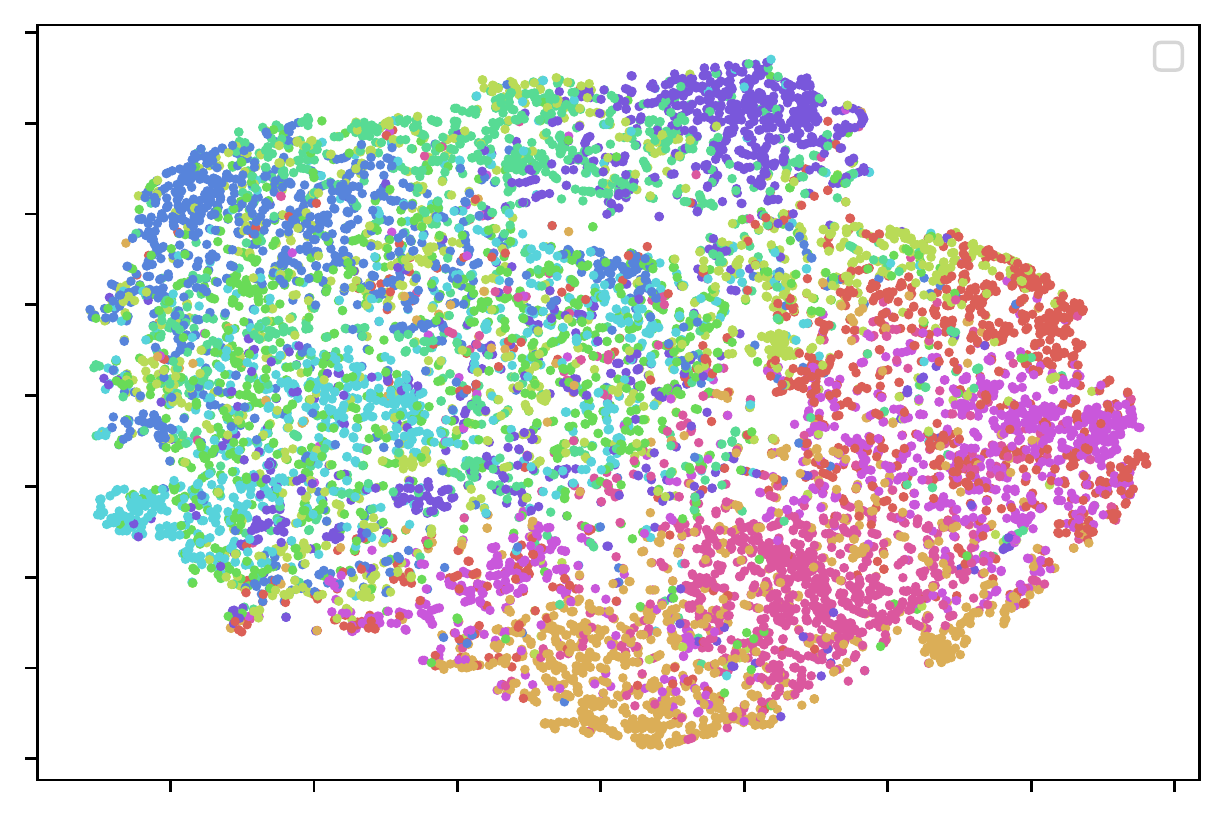}}
  \subfigure[Round 50]{\label{fig:online-online}\includegraphics[width=30mm]{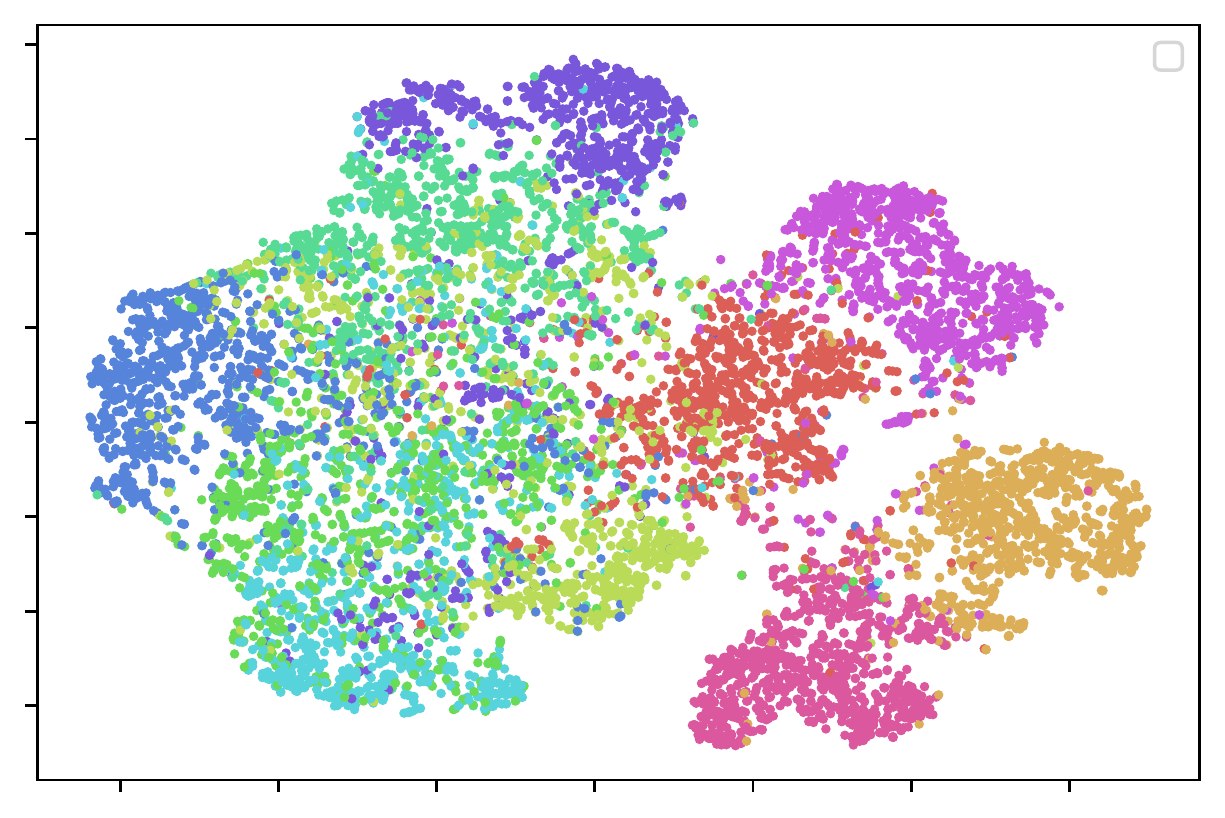}}
  \subfigure[Round 100]{\label{fig:tsne-fedu}\includegraphics[width=30mm]{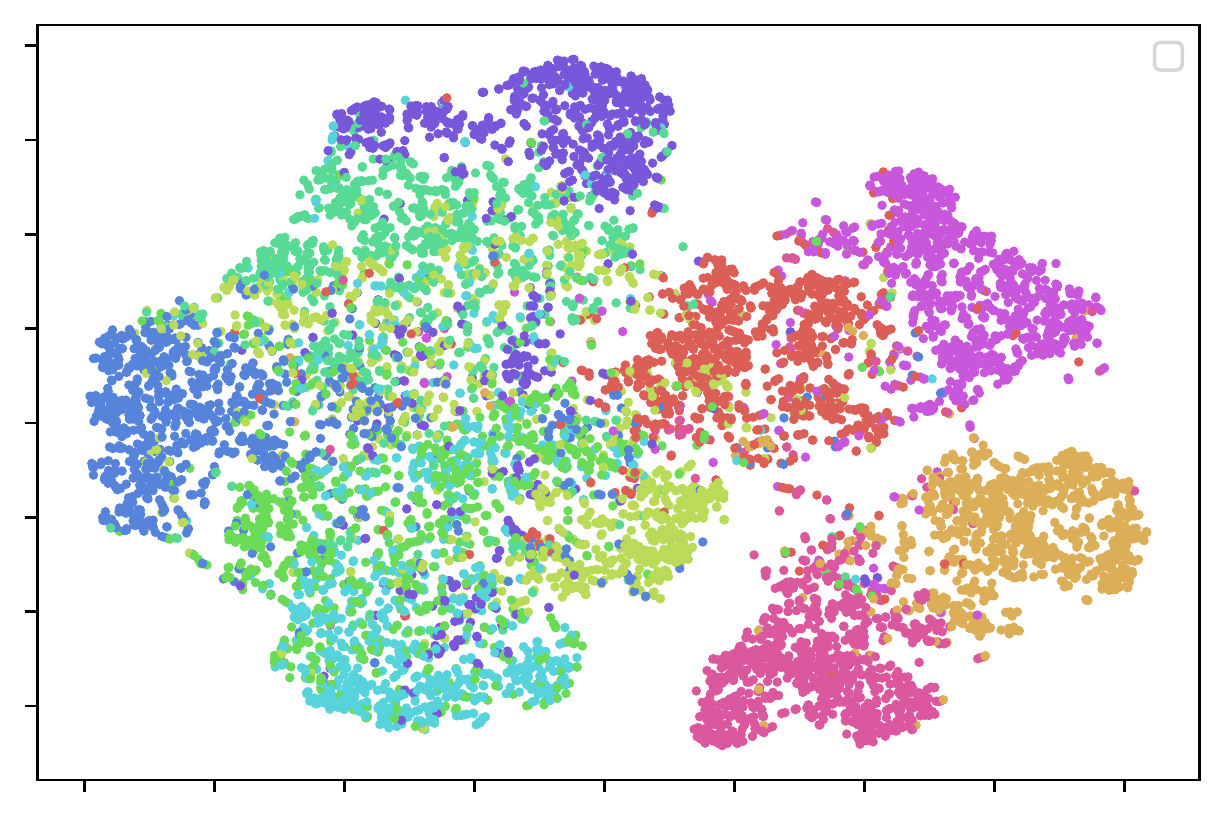}}
\end{center}
  \caption{t-SNE visualization of feature representations learned by FedMCC on CIFAR-10. Different colors mean different clusters}
  \label{fig:t-SNE-CIFAR-10-iid}
\end{figure*}

For CIFAR-10 and IID data, we visualize the t-SNE embeddings produced by the cluster projector during the communication rounds in Fig.~\ref{fig:t-SNE-CIFAR-10-iid}. 
FedMCC progressively trains the embedding from indistinguishable at the beginning to clear at the end.  


\begin{table*}[ht]
    \centering
    \caption{Clustering accuracy  (\%) on IID datasets}
    \label{tab:federated_iid}
    \resizebox{\linewidth}{!}{
    \begin{tabular}{l*{15}{r}}
      \toprule
      Dataset 
      &\multicolumn{3}{c}{\textbf{CIFAR-10}}
      &\multicolumn{3}{c}{\textbf{CIFAR-100}}
      &\multicolumn{3}{c}{\textbf{STL-10}}
      &\multicolumn{3}{c}{\textbf{ImageNet-10}}
      &\multicolumn{3}{c}{\textbf{MNIST}}\\
      \cline{2-16}
      Method 
      & NMI & ACC & ARI 
      & NMI & ACC & ARI 
      & NMI & ACC & ARI
      & NMI & ACC & ARI 
      & NMI & ACC & ARI \\
      \midrule
      FedU + K-means
      & 47.0 & 57.6  & 36.1
      & 37.0 & 21.9 & 11.5
      & 35.2 & 39.4  & 20.2
      & 49.9& 61.8 & 37.8
      & 77.4 & 80.6 & 71.2 \\
      FedU + MCC
      & 52.9 & 62.5 & 43.9
      &31.7 & 16.7 & 6.9
      & 35.2 & 41.8 & 21.9
      &53.7 & 63.1 & 44.9
      & 79.9 & 83.7 & 75.2 \\
      Fed + CC
      & 54.6 & 64.2 & 45.4
      & 28.7 & 14.9 & 6.9
      & 41.8 & 49.3 & 30.7
      & 53.3 & 63.6 & 50.4
      & 75.0 & 80.8 & 68.2 \\
      \midrule
      FedMCC (ours)
      & 57.0 & \bf 69.4 & 50.6 
      & 35.6 & \bf 22.0 & 11.8
      &41.7 & \bf 49.6 & 30.8
      & 61.6 & \bf 67.3 & 54.1
      & 79.6 & \bf 87.3 & 76.2 \\
     \midrule
      NCC~\cite{NCC-21-huang} (Centralized)
      & 88.6 & 94.3 & 88.4 
      & 60.6 & 61.4 & 45.1
      & 75.8 & 86.7 & 73.7 
      & 89.6 & 95.6 & 90.6 
      & - & - & -  \\
    \bottomrule
  \end{tabular}}
\end{table*}

\begin{table*}[ht]
    \centering
    \caption{Clustering accuracy (\%) on non-IID datasets}
    \label{tab:federated_noniid}
    \resizebox{\linewidth}{!}{
    \begin{tabular}{l*{15}{r}}
      \toprule
      Dataset 
      &\multicolumn{3}{c}{\textbf{CIFAR-10}}
      &\multicolumn{3}{c}{\textbf{CIFAR-100}}
      &\multicolumn{3}{c}{\textbf{STL-10}}
      &\multicolumn{3}{c}{\textbf{ImageNet-10}}
      &\multicolumn{3}{c}{\textbf{MNIST}}\\
      \cline{2-16}
      Method 
      & NMI & ACC & ARI 
      & NMI & ACC & ARI 
      & NMI & ACC & ARI
      & NMI & ACC & ARI 
      & NMI & ACC & ARI \\
      \midrule
      FedU + K-means
      &  45.8& 48.7 & 30.0 
      & 37.0 & 21.9 & 11.5
      &  34.8 & 38.9 & 19.9
      &40.0& 48.0 & 29.3
      & 42.0 & 41.8 & 25.8\\
      FedU + MCC
      & 38.2 & 47.2 & 26.6
      &31.7 &16.7 &6.9
      & 33.6 & 36.9 &19.6
      &39.8 & 47.6 & 29.0
     & 53.1 & 55.1 & 38.2\\
      Fed + CC
      & 38.4 & 43.4 & 25.6
      & 30.5 & 14.7 & 5.8
      & 34.2 & 38.7 & 20.1
      & 44.3 & 50.4 & 32.6
      & 55.6 & 54.5 & 40.4 \\
      \midrule
      FedMCC (ours)
      & 45.5 & \textbf{49.5} &  29.8
      & 37.9 & \textbf{22.6} & 11.4 
      & 34.3 & \bf 43.0 & 21.9
      &44.8 & \bf 51.1 & 33.2
      & 59.4 & \textbf{58.0} & 44.4 \\
      \midrule
      NCC~\cite{NCC-21-huang} (Centralized)
      & 88.6 & 94.3 & 88.4 
      & 60.6 & 61.4& 45.1
      & 75.8 & 86.7 & 73.7 
      & 89.6 & 95.6 & 90.6 
      & - & - & - \\
    \bottomrule
  \end{tabular}}
\end{table*}

\subsection{Linear and Semi-Supervised Learning Evaluation}
Even though our focus in this paper is training a clustering scheme in the federated setting, we assess the performance of FedMCC on linear evaluation and semi-supervised setups to prove the effectiveness of the learned representations.

\begin{table}[!ht]
    \centering
\begin{minipage}[t]{0.48\linewidth}\centering
\caption{Linear evaluation on Resnet-50}\vspace{-5pt}
\scalebox{0.85}{\begin{tabular}{l*{6}{r}}
\toprule
Dataset &
\multicolumn{2}{c}{\textbf{CIFAR-10}} &
\multicolumn{1}{c}{} &
\multicolumn{2}{c}{\textbf{CIFAR-100}}
\\ 
\cline{2-6}
Method
 & IID & Non-IID & & IID & Non-IID
\\
\hline 
Single client training\cite{fedu-21-zhuang} 
& 83.2 & 77.8
& & 57.2 & 55.2 \\
FedSimCLR~\cite{simclr-20-chen, FedCA-20-zhang} 
& 68.1 & 64.1 
& & 39.8 & 38.7 \\ 
FedCA~\cite{FedCA-20-zhang} 
& 71.3 & 68.0 
& & 43.3 & 42.3 \\ 
FedSimSiam~\cite{simsiam-21-chen, fedu-21-zhuang} 
& 79.6 & 76.7 
& & 46.3 & 48.8 \\
FedU~\cite{fedu-21-zhuang} 
& 86.5 & 83.2
& &59.5 & 61.9 \\

FedEMA~\cite{fedema-2022} ($\lambda = 0.8$) 
& 86.1 & \bf 85.3
& &\bf 60.9 & 62.5 \\
FedEMA~\cite{fedema-2022} ($\tau = 0.7$) 
& 85.1 & 84.3
& &59.5 & \bf  62.8 \\ \hline
FedMCC 
& \bf 86.8 & \bf 85.3
& & 60.8  & 62.3 \\
\hline
\label{tab:res50-linear-eval}
\end{tabular}}\vspace{-10pt}
\end{minipage}\hfill%
\begin{minipage}[t]{0.48\linewidth}\centering
\caption{Linear evaluation on Resnet-18}\vspace{-5pt}
\scalebox{0.85}{\begin{tabular}{l*{4}{r}}
\toprule
Dataset &
\multicolumn{2}{c}{\textbf{CIFAR-10}} &
\multicolumn{2}{c}{\textbf{CIFAR-100}}
\\ 
\cline{2-5}
Method
 & IID & Non-IID & IID & Non-IID
\\
\hline 
Single client
& 81.2 & 72.0 
& 51.3 & 49.7 \\
FedU 
& 85.2 & 78.7
& 56.5 & 57.1 \\ 
FedEMA ($\lambda = 0.8$)
& 85.6 & 82.8 
& 57.9 & 61.2 \\ 
FedEMA ($\tau = 0.7$) 
&  86.3 &  83.3 
&  58.6 & \bf 61.8 \\ 
\hline
FedMCC 
& 85.5 & 82.7
& 57.2 & 57.1 \\ 
FedMCC (Tuned)
& \bf 87.8 & \bf 85.2
& \bf 59.6 & 59.9 \\
\hline\label{tab:res18-linear-eval}
\end{tabular}}
\end{minipage}\hfill
\end{table}

\textbf{Linear Evaluation:} We train a linear classifier on top of the frozen representation trained by FedMCC and measure the Top 1\% accuracy.
Tables \ref{tab:res50-linear-eval} and \ref{tab:res18-linear-eval} compare our schemes with previous approaches~\cite{fedu-21-zhuang,FedCA-20-zhang} by using different backbones. With a ResNet-50, FedMCC consistently outperforms FedEMA~\cite{fedema-2022} all existing schemes on CIFAR10 dataset. The performance of FedMCC is still very close to FedEMA on  CIFAR-100. Note that FedEMA utilizes mechanisms for tuning the momentum parameter $m$ in the learning phase. If we also optimize $m$, we can push the accuracies further, as demonstrated in the results for ResNet-18. Although the untuned FedMCC with $m=0.99$ falls short of FedEMA, by tuning $m=0.988$, we can outperform FedEMA in most scenarios. In particular, with a ResNet-18, FedMCC obtains 85.2\% top-1 accuracy on Non-IID CIFAR-10 dataset, which is almost 2\% higher than the closest competitor.
 
\begin{table}
\caption{Top-1 accuracy (\%) comparison under the semi-supervised protocol}
    \begin{center}\scalebox{1.0}{
\begin{tabular}{l*{5}{r}}
\toprule
Dataset &
\multicolumn{2}{c}{\textbf{1\%}} &
\multicolumn{2}{c}{\textbf{10\%}}
\\ 
\cline{2-3} \cline{4-5}
Method
 & IID & \scriptsize Non-IID & IID & \scriptsize Non-IID
\\
\hline 
Single client  (Res18)\cite{fedu-21-zhuang}
& 74.8 & 60.3
& 78.1 & 70.6 \\
Single client (Res50)\cite{fedu-21-zhuang}
& 74.8 & 63.7
& 80.3 & 74.3 \\
FedSimCLR (Res50)~\cite{simclr-20-chen, FedCA-20-zhang}
& 50.0 & 26.0 
& 60.7 & 33.8 \\ 
FedCA (Res50)~\cite{FedCA-20-zhang}
& 50.7 & 28.5 
& 61.0 & 36.3 \\ 
FedU (Res50)~\cite{fedu-21-zhuang}    
& 79.4 & 71.2
& \textbf{83.1} & 80.1 \\ 
FedU (Res18)~\cite{fedu-21-zhuang}    
& 79.4 & 68.3
& 82.6 & 78.5 \\ 
FedEMA ($\lambda=1$, Res18) & - &  72.8 & - & 79.0 
\\ 
\hline
FedMCC (Res18)
& 80.0 & 72.2 
& 82.9 & \textbf{82.0} \\ 
FedMCC (Res18) (Tuned)
& \bf 82.9 & \bf 73.3
& \bf 83.1 & \textbf{82.0} \\ 
\hline
\end{tabular}}
\end{center}
\label{tab:semi-sup}
\end{table}

\textbf{Semi-Supervised Learning:} We follow the same experimental procedure in \cite{semi-19-zhai, simclr-20-chen, fedu-21-zhuang} by using the labels of 1\% or 10\% of the samples during training on the CIFAR-10 dataset. 
Table \ref{tab:semi-sup} reveals that FedMCC improves over existing methods, especially for the non-IID setting, obtaining 82.0\% top-1 accuracy for $10\%$ labeled data with or without tuning the decay rate $m$. The tuned FedMCCs all use $m=0.988$ as before.

\subsection{Centralized Clustering Evaluation}
We compare our MCC method with previous methods in Table \ref{tab:centralized results}. MCC achieves the state-of-the art clustering accuracy on benchmark datasets including STL-10 with 87.0\% and ImageNet-10 with 95.9\%, showing the advantages of our proposed clustering scheme in centralized settings as well. Our MCC scheme also improves the performance of the CC scheme on all datasets. As compared with the best available methods, the performance of MCC falls short in certain datasets such as the CIFAR-10 and Tiny-Imagenet. The specific nature of the datasets that makes MCC performs well (or worse) remains an interesting avenue for further research. 

\begin{figure}
\begin{center}
  \subfigure[\scalebox{1}{Epoch 0 (NMI = 0.211)}]{\label{fig:centra-STL-10-epo0}\includegraphics[width=50mm]{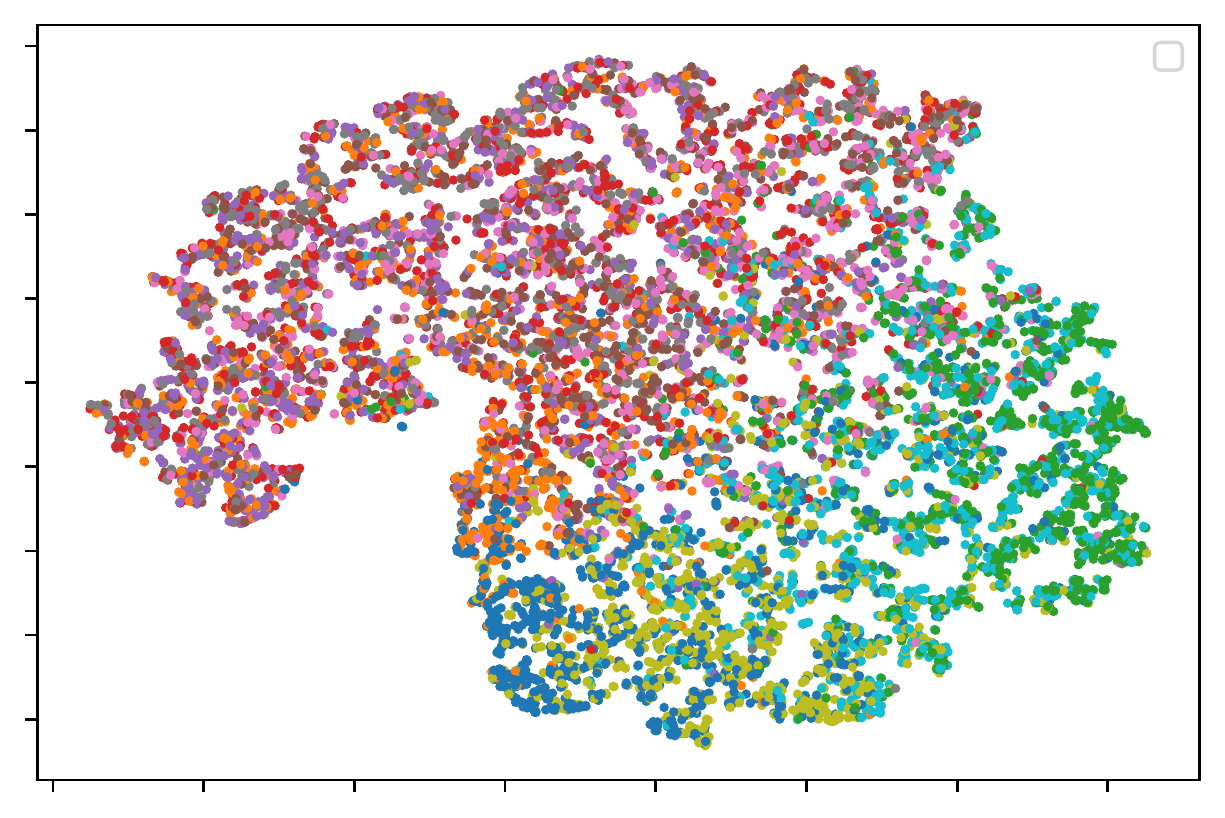}}
  \subfigure[\scalebox{1}{Epoch 500 (NMI = 0.706)}]{\label{fig:centra-STL-10-epo100}\includegraphics[width=50mm]{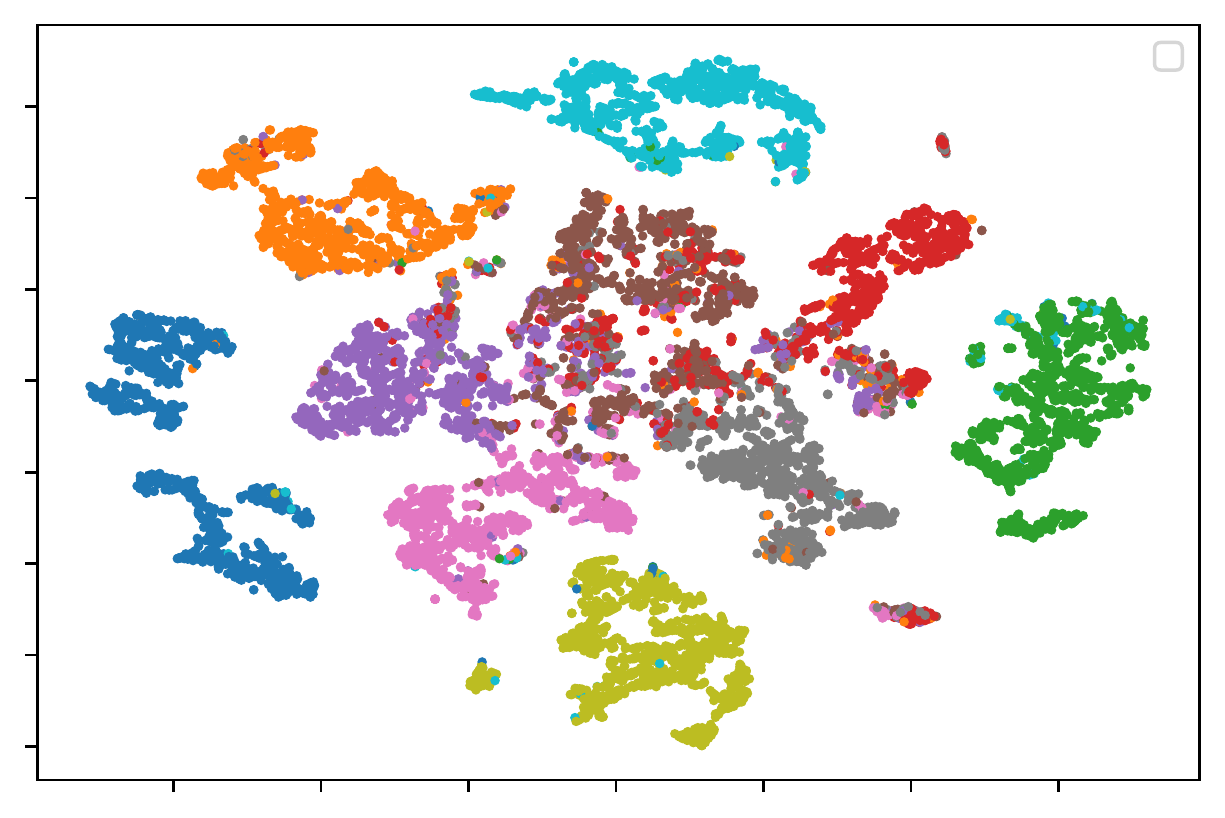}}
\end{center}
  \caption{t-SNE plots of MCC for STL-10 dataset }
  \label{fig:centra-t-SNE-STL-10}
\end{figure}

We also provide the t-SNE plots on the STL-10 dataset in Fig~\ref{fig:centra-t-SNE-STL-10}. Different colors denote the different predicted labels from the cluster-level projector. 
After 500 epochs, MCC produces clear boundaries for cluster assignments.



\newsavebox\Dogs
\newlength\wdDogs

\sbox{\Dogs}{\textbf{ImageNet-Do}}
\setlength{\wdDogs}{\wd\Dogs}


\begin{table*}[ht]
    \centering
    \caption{Clustering performance of different centralized algorithms}
    \label{tab:centralized results}
    \resizebox{\textwidth}{!}{
    \begin{tabular}{l*{18}{>{\centering\arraybackslash}p{0.3333\wdDogs}}}
      \toprule
      Dataset 
      &\multicolumn{3}{c}{\textbf{CIFAR-10}}
      &\multicolumn{3}{c}{\textbf{CIFAR-20}}
      &\multicolumn{3}{c}{\textbf{STL-10}}
      &\multicolumn{3}{c}{\textbf{ImageNet-10}}
      &\multicolumn{3}{c}{\textbf{ImageNet-Dogs}}
      &\multicolumn{3}{c}{\textbf{Tiny-ImageNet}}\\
      \cline{2-19}
      Method 
      & NMI & ACC & ARI 
      & NMI & ACC & ARI 
      & NMI & ACC & ARI 
      & NMI & ACC & ARI 
      & NMI & ACC & ARI 
      & NMI & ACC & ARI \\
      \midrule
      k-means 
      & 8.7 & 22.9 & 4.9    
      & 8.4  & 13.0 & 2.8
      & 12.5 & 19.2 & 6.1  
      & 11.9 & 24.1 & 5.7
      & 5.5 & 10.5 & 2.0
      & 6.5 & 2.5 & 0.5 \\
      SC 
      & 10.3 & 24.7 & 8.5
      & 9.0 & 13.6 & 2.2 
      & 9.8 & 15.9 & 4.8
      & 15.1 & 27.4 & 7.6
      & 3.8 & 11.1 & 1.3
      & 6.3 & 2.2 & 0.4\\
      AE 
      & 23.9 & 31.4 & 16.9 
      & 10.0 & 16.5 & 4.8 
      & 25.0 & 30.3 & 16.1 
      & 21.0 & 31.7 & 15.2
      & 10.4 & 18.5 & 7.3
      & 6.9 & 2.7 & 0.5\\
      VAE 
      & 24.5 & 29.1 & 16.7 
      & 10.8 & 15.2 & 4.0
      & 20.0 & 28.2 & 14.6 
      & 19.3 & 33.4 & 16.8
      & 10.7 & 17.9 & 7.9
      & 11.3 & 3.6 & 0.6\\
      JULE 
      & 19.2 & 27.2 & 13.8 
      & 10.3 & 13.7 & 3.3
      & 18.2 & 27.7 & 16.4 
      & 17.5 & 30.0 & 13.
      & 5.4 & 13.8 & 2.8
      & 10.2 & 3.3 & 0.6\\
      DEC 
      & 25.7 & 30.1 & 16.1 
      & 13.6 & 18.5 & 5.0
      & 27.6 & 35.9 & 18.6 
      & 28.2 & 38.1 & 20.3
      & 12.2 & 19.5 & 7.9
      & 11.5 & 3.7 & 0.7\\
      DAC 
      & 39.6 & 52.2 & 30.6 
      & 18.5 & 23.8 & 8.8
      & 36.6 & 47.0 & 25.7 
      & 39.4 & 52.7 & 30.2
      & 21.9 & 27.5 & 11.1
      & 19.0 & 6.6 & 1.7\\
      IIC 
      & 51.3 & 61.7 & 41.1 
      & -    & 25.7 & -      
      & 43.1 & 49.9 & 29.5  
      & -    & -    & -   
      & -    & -    & -
      & -     & -     & - \\    
      DCCM 
      & 49.6 & 62.3 & 40.8 
      & 28.5 & 32.7 & 17.3
      & 37.6 & 48.2 & 26.2 
      & 60.8 & 71.0 & 55.5
      & 32.1 & 38.3 & 18.2
      & 22.4 & 10.8 & 3.8\\
      PICA 
      & 56.1 & 64.5 & 46.7    
      & 29.6 & 32.2 & 15.9
      & -    & -    & -     
      & 78.2 & 85.0 & 73.3
      & 33.6 & 32.4 & 17.9
      & 27.7 & 9.8 & 4.0\\     
      CC 
      &  70.5 & 79.0 & 63.7 
      & 43.1 & 42.9 & 26.6
      & 76.4 & 85.0 & 72.6 
      & 85.9 & 89.3 & 82.2
      & 44.5 & 42.9 & 27.4
      & 34.0 & 14.0 & 7.1\\
      SCAN 
      &  79.7 & 88.3 & 77.2 
      & 48.6 & 50.7 & 33.3
      & 80.9 & 69.8 & 64.6 
      & - & - & - 
      & - & - & -
       & - & - & -\\
      GCC 
      & 76.4 & 85.6 & 72.8 
      & 47.2 & 47.2 & 30.5
      & 68.4 & 78.8 & 63.1 
      & 84.2 & 90.1 & 82.2
      & 49.0 & 52.6 & 36.2
      &34.7 & 13.8 & 7.5\\
      MiCE 
      & 73.7& 83.5& 69.8 
      & 43.6 & 44.0& 28.0
      & 63.5 & 75.2 & 57.5 
      & - & - & - 
      & 42.3 & 43.9 & 28.6
      &- & - & -\\
      IDFD 
      & 71.1& 81.5& 66.3 
      & 42.6 & 42.5& 26.4 
      & 64.3 & 75.6 & 57.5 
      & 89.8 & 95.4 & 90.1
      & 54.6 & 59.1 & 41.3
      &- & - & -\\
      PCL 
      & 80.2 & 87.4 & 76.6 
      & 52.8 & 52.6 & 36.3
      & 71.8 & 41.0 & 67.0
      & 84.1 & 90.7 & 82.2
      & 44.0 & 41.2 & 29.9
      &- & - & -\\
      NCC~\cite{NCC-21-huang}
      & \bf 88.6 & \bf 94.3 & \bf 88.4 
      & \bf 60.6 & \bf 61.4& \bf 45.1
      & 75.8 & 86.7 & 73.7 
      & 89.6 & 95.6 & 90.6 
      & \bf 69.2 & \bf 74.5 & \bf 62.7
      & \bf 40.5 & \bf 25.6 & \bf 14.3\\
      \midrule
      MCC (ours) 
      & 76.2 & 84.9 & 72.7
      & 44.0 & 45.4 & 28.5
      & \textbf{77.2} & \textbf{87.0} & \textbf{74.8}
      & \textbf{90.4} & \textbf{95.9} & \textbf{91.1}
      & 53.0 & 54.7 & 39.5
      & 35.09 & 15.2 & 7.5 \\
    \bottomrule
  \end{tabular}}
\end{table*}

%% file: 05-ablation.tex
\section{Ablation Studies}
\label{secablation}

{\bf Effect of Target Network:} To prove the effectiveness of the target network, we remove the exponential moving average of the online networks. 
Tables \ref{tab:federated_iid} and \ref{tab:federated_noniid}  show the performance of the resulting Fed + CC scheme, described in Section \ref{fedclusteval}. 
Without the target networks, the performance decreases notably from FedMCC. A similar conclusion can be made from the results for the centralized scenario shown in Table~\ref{tab:centralized results}, through observing the relative improvement of MCC over CC. 

{\bf Impact of Four Representations:} We recall that FedU and variants feed the two independent feature vectors to the online and target networks. On the other hand, FedMCC implements the novel idea of feeding both vectors to both networks, resulting in four representations. We can measure the resulting performance gains by comparing FedMCC with FedU + MCC. Table~\ref{tab:centralized results} reveals that the gains are within the range 6-7\%.


 \begin{figure}
\begin{center}
  \subfigure[MNIST-IID]{\label{fig:commu-mnist-iid}\includegraphics[width=50mm]{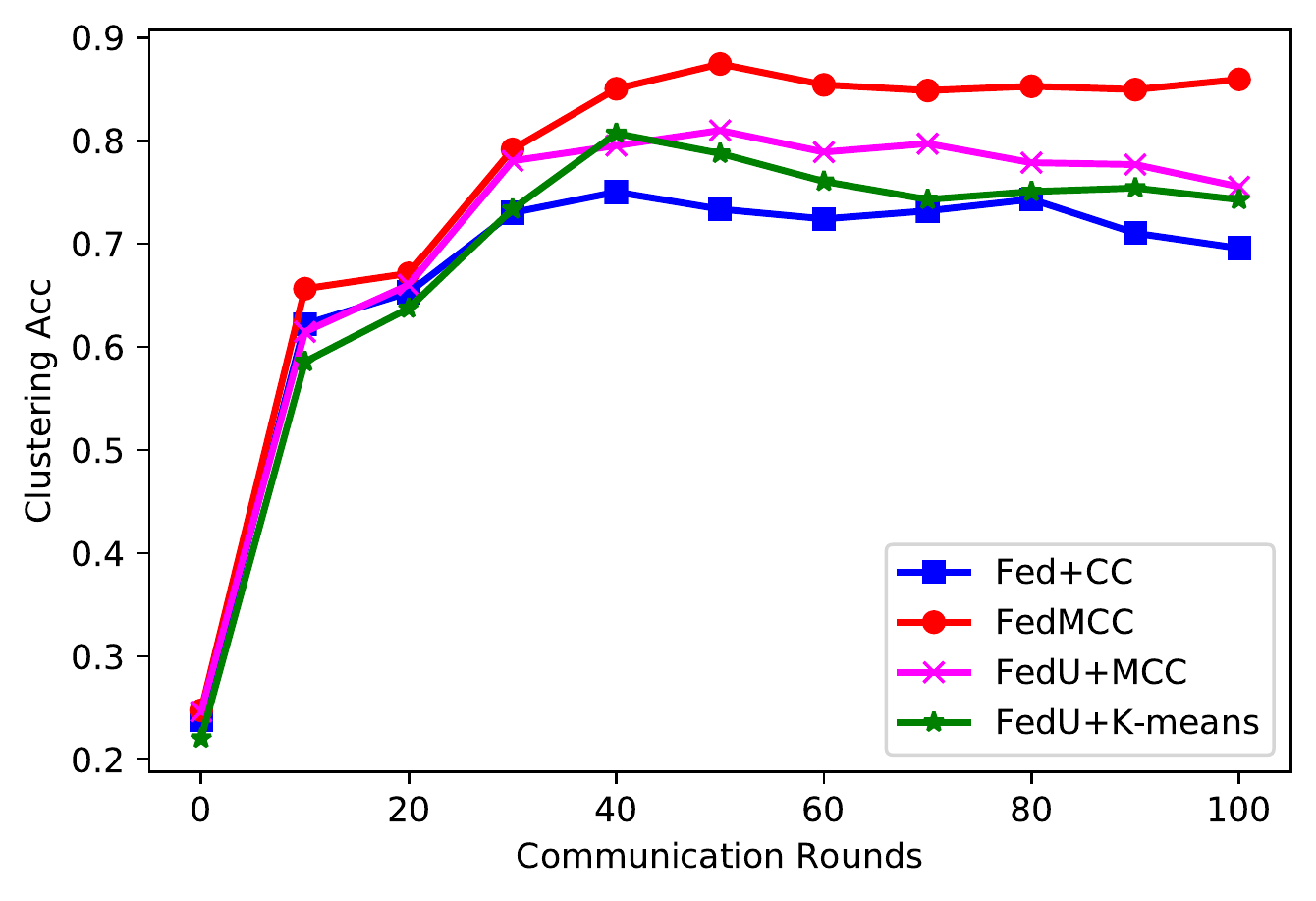}}
  \subfigure[MNIST-NonIID]{\label{fig:commu-mnist-noniid}\includegraphics[width=50mm]{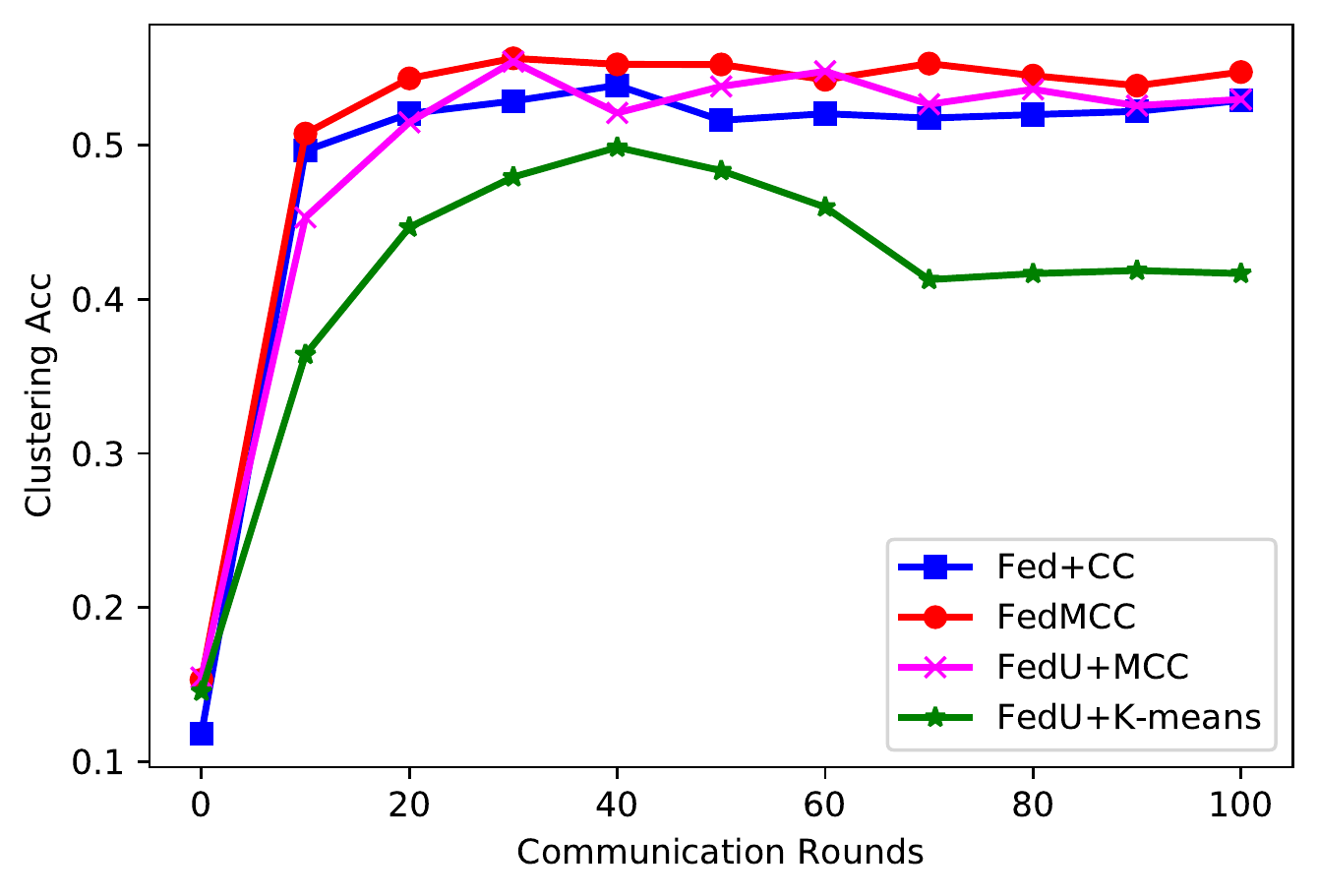}}
\end{center}
  \caption{Clustering accuracy versus communication rounds}
  \label{fig:ablation}
\end{figure}

{\bf Communication Rounds:} Fig.~\ref{fig:ablation} shows the clustering accuracy with respect to communication rounds for the MNIST dataset. 
For IID distribution of data, FedCC and FedU + K-means  have similar performances, and FedMCC achieves the best performance at all rounds. In the Non-IID setting, FedMCC and FedCC perform better than FedU + K-means at each communication round.

{\bf Target Network Aggregation: } We also conduct experiments by updating andaggregating both online and target networks during the communication round. 
Table~\ref{tab:update-online-target-results} 
shows the results when considering the target networks in communication protocol. 
We used the backbone ResNet-18 for this ablation study. Compared to the results in Tables \ref{tab:federated_iid} and \ref{tab:federated_noniid}, there is a major loss in performance in all scenarios. 

\begin{table}
    \centering
    \caption{Updating online \& target nets}
    \label{tab:update-online-target-results}\scalebox{1}{
    \begin{tabular}{l*{3}{r}}
      \toprule
      Method
      &\multicolumn{3}{c}{Update Both}\\
      \cline{2-4}
      Dataset 
      & NMI & ACC & ARI \\
      \midrule
      \textbf{CIFAR-10} (IID)
      & 53.2 & 58.3 & 43.5\\
      \textbf{CIFAR-10} (Non-IID)
      & 34.9 & 40.4 & 22.8 \\
      \textbf{CIFAR-100} (IID)
      & 33.0 & 18.3 & 7.6 \\
      \textbf{CIFAR-100} (Non-IID)
      & 30.9 & 15.5 & 6.4\\
      \textbf{MNIST} (IID)
      & 62.6 & 62.9 & 47.7\\
      \textbf{MNIST} (Non-IID)
      & 24.5& 37.3 & 14.9\\
    \bottomrule
  \end{tabular}}
\end{table}

\textbf{Effect of Local Epochs:} In Table \ref{tab:effect_of_local_epochs}, we consider $500$ total epochs for each client. We vary local epochs $E = 1, 2, 5, 10$ with the corresponding number of communication rounds $R = 500, 250, 100, 50$. We can observe that  both the number of local epochs and communication rounds are important, and there is a non-trivial optimal operating point.

\textbf{Impact of Momentum Parameter:} We have already seen through the results in  Table \ref{tab:res18-linear-eval} and \ref{tab:semi-sup} that optimizing the momentum parameter $m$ can improve the performance. Here, we show results for the choice $m=0.996$ in Table \ref{tab:effect_of_momentum_parameter}, considered in the original BYOL paper  \cite{byol-20}. The performance becomes noticeably worse, suggesting a strong dependence of the performance on $m$. Optimization of the decay rate for BYOL-type algorithms in the federated setting remains a very interesting avenue for further research. We have generally not performed decay rate optimization for fair comparison with existing studies (most previous work such as FedU also consider $m=0.99$).

\newsavebox\CTI
\newsavebox\CTN
\newsavebox\CHI
\newsavebox\CHN

\newlength\wdCTI
\newlength\wdCTN
\newlength\wdCHI
\newlength\wdCHN

\sbox{\CTI}{\textbf{CIFAR-10} (IID)}
\setlength{\wdCTI}{\wd\CTI}

\sbox{\CTN}{\textbf{CIFAR-10} (Non-IID)}
\setlength{\wdCTN}{\wd\CTN}

\sbox{\CHI}{\textbf{CIFAR-100} (IID)}
\setlength{\wdCHI}{\wd\CHI}

\sbox{\CHN}{\textbf{CIFAR-100} (Non-IID)}
\setlength{\wdCHN}{\wd\CHN}

\begin{table*}[ht]
    \centering
    \caption{Effect of local epochs}\vspace{-5pt}
    \label{tab:effect_of_local_epochs}
    \resizebox{\textwidth}{!}{
    \begin{tabular}{l*{3}{>{\centering\arraybackslash}p{0.333\wdCTI}}*{3}{>{\centering\arraybackslash}p{0.333\wdCTN}}*{3}{>{\centering\arraybackslash}p{0.333\wdCHI}}*{3}{>{\centering\arraybackslash}p{0.333\wdCHN}}}
      \toprule
      Dataset 
      &\multicolumn{3}{c}{\usebox{\CTI}}
      &\multicolumn{3}{c}{\usebox{\CTN})}
      &\multicolumn{3}{c}{\usebox{\CHI}}
      &\multicolumn{3}{c}{\usebox{\CHN}}\\
      \cline{2-13}
      Method 
      & NMI & ACC & ARI 
      & NMI & ACC & ARI 
      & NMI & ACC & ARI 
      & NMI & ACC & ARI \\
      \midrule
      $E = 1, R = 500$
      & 48.7 & 57.6& 36.2
      & 39.5 & 40.9 & 25.9
      & 34.6 & 19.2 & 8.4
      & 32.3 & 17.1 & 7.1 \\
      $E = 2, R = 250$
      & 50.5 & 59.1 & 39.3
      & 41.4 & 45.0 &  28.4
      & 34.4& 19.5 & 8.27
      & 32.9 & 17.6& 3.4\\
      $E = 5, R = 100$
      & 57.0 & 69.4 & 50.6
      & 45.5 & 49.5 &  29.8
      & 35.6 & 22.0 & 11.8
      & 37.9 & 22.6 & 11.4 \\
      $E = 10, R = 50$
      &50.7 & 58.7 & 41.3
      & 37.4 & 43.1 & 24.2
      &34.3& 19.0 & 8.2
      & 24.1 & 12.1 & 4.2\\
    \bottomrule
  \end{tabular}}
\end{table*}

\begin{table*}[ht]
    \centering
    \caption{Effect of the momentum parameter}\vspace{-5pt}
    \label{tab:effect_of_momentum_parameter}
    \resizebox{\textwidth}{!}{
    \begin{tabular}{l*{3}{>{\centering\arraybackslash}p{0.333\wdCTI}}*{3}{>{\centering\arraybackslash}p{0.333\wdCTN}}*{3}{>{\centering\arraybackslash}p{0.333\wdCHI}}*{3}{>{\centering\arraybackslash}p{0.333\wdCHN}}}
      \toprule
      Dataset 
      &\multicolumn{3}{c}{\textbf{CIFAR-10} (IID)}
      &\multicolumn{3}{c}{\textbf{CIFAR-10} (Non-IID)}
      &\multicolumn{3}{c}{\textbf{CIFAR-100} (IID)}
      &\multicolumn{3}{c}{\textbf{CIFAR-100} (Non-IID)}\\
      \cline{2-13}
      Method 
      & NMI & ACC & ARI 
      & NMI & ACC & ARI 
      & NMI & ACC & ARI 
      & NMI & ACC & ARI \\
      \midrule

      FedMCC ($m = 0.996$)
      & 45.6 & 52.4& 34.0
      & 37.5 & 45.2 &  25.7
      & 33.9 & 19.0 & 7.8
      & 31.8& 16.6 &6.6 \\
      FedMCC ($m = 0.99$)
      & \bf 57.0 & \bf 69.4 & \bf 50.6
      & \bf 45.5 & \bf 49.5 &  \bf 29.8
      & \bf 35.6 & \bf 22.0 &\bf 11.8
      & \bf 37.9 &\bf  22.6 & \bf 11.4 \\
    \bottomrule
  \end{tabular}}
\end{table*}




%% file: 035-memefficientsupp.tex
\section{Memory-Efficient MCC and FedMCC}
\label{secmemefficient}
An important special case of the federated learning paradigm is when the clients are low-cost edge devices that may have memory limitations. The amount of training memory needed for a naive implementation of the gradient descent updates for the loss function in (\ref{mcclossfunc}) or its federated counterparts in (\ref{eq:fed_Lins}) and (\ref{eq:fed_Lclu}) is linear in the batch size $n$. For the example case of the CC, this is because the network outputs $\{z_i^{\sigma,\nu}, y_i^{\sigma,\nu}\},\,i=1,\ldots,n$ for different members of a batch $x_i,\,i=1,\ldots,n$ are coupled in a non-linear manner in the loss functions. Hence, to perform a gradient update, the size of the computation graph should be roughly of size $n\eta$, where $\eta$ represents the neural network size. This is in contrast to ordinary batch loss of the form $\sum_{i=1}^n \ell(x_i)$, where $x_i$ are inputs and $\ell$ is some loss function. In this case, the gradient of the entire loss function can be calculated through accumulation using only one computation graph of size $\eta$.


Our idea for constructing memory-efficient MCC and FedMCC is akin to the scheme presented in \cite{gao2021scaling} for the case of ordinary contrastive learning. We first calculate the gradients of the loss function with respect to the network outputs or representations. These gradients are calculated over one pass over the batch and stored on the memory. The memory overhead of this first stage is relatively low as the dimensionality of representation vectors is much smaller than that of the parameter space. A second pass over the batch can then update the network parameters. To describe the details of this procedure, let us now extend the CC scheme to a memory-efficient CC scheme. Extensions to MCC and FedMCC can be accomplished in the same fashion.

We recall the loss function for the CC scheme in (\ref{lccloss}). The goal is to calculate the gradient $\frac{\partial L_{CC}}{\partial \theta}$, where $\theta$ represents the neural network parameters. For each $i\in\{1,\ldots,n\}$, let us define the gradients
\begin{align}
\label{alphas123} \alpha_{i,1} \triangleq \frac{1}{2}\frac{\partial L(\mathbf{z}^a, \mathbf{z}^b; \tau_I)}{\partial z_i^a}, \alpha_{i,2} \triangleq \frac{1}{2}\frac{\partial L(\mathbf{z}^a, \mathbf{z}^b; \tau_I)}{\partial z_i^b},  \alpha_{i,3} \triangleq \frac{1}{2}\frac{\partial L(\mathbf{c}^a, \mathbf{c}^b; \tau_C)}{\partial y_i^a}, \\ \label{alphas456} \alpha_{i,4} \triangleq \frac{1}{2}\frac{\partial L(\mathbf{c}^a, \mathbf{c}^b; \tau_C)}{\partial y_i^b},
 \alpha_{i,5} \triangleq \frac{\partial H(\mathbf{c}^a)}{\partial y_i^a}, \alpha_{i,6} \triangleq \frac{\partial H(\mathbf{c}^b)}{\partial y_i^b}.
\end{align}
The partial derivative of the overall loss function with respect to one of the parameters $\theta'\in\theta$ is then expressed as 
\begin{align}
\label{lccform1}
\frac{\partial L_{CC}}{\partial \theta} = \sum_{i=1}^n \Bigl[ \Bigl\langle \alpha_{i,1},\frac{\partial z_i^a}{\partial \theta'}\Bigr\rangle + \Bigl\langle \alpha_{i,2},\frac{\partial z_i^b}{\partial \theta'}\Bigr\rangle + \Bigl\langle \alpha_{i,3} \!+\! \alpha_{i,5},\frac{\partial y_i^a}{\partial \theta'}\Bigr\rangle \!+\! \Bigl\langle \alpha_{i,4}\!+\!\alpha_{i,6},\frac{\partial y_i^b}{\partial \theta'}\Bigr\rangle \Bigl].
\end{align}
Note that, for any input index $i\in\{1,\ldots,n\}$ in a batch, the gradients $\frac{\partial z_i^{\sigma}}{\partial \theta'},\,\frac{\partial y_i^{\sigma}}{\partial \theta'},\,\sigma\in\{a,b\}$ in (\ref{lccform1}) depend only on $x_i$ (and are independent of $x_j,\,j\neq i$). Hence, if $\alpha_{i,j},\,i\in\{1,\ldots,n\},\,j\in\{1,\ldots,6\}$ are known, or calculated beforehand, the entire partial derivative (\ref{lccform1}) assumes the form of an ``ordinary batch loss'' $\sum_{i=1}^n \ell(x_i)$, as described in the beginning of this section. In fact, the parameters $\alpha_{i,j}$ can simply be calculated by a single forward pass of the entire batch (possibly one input at a time to save memory) before the gradient updates. As a result, once $\alpha_{i,j}$ are known, (\ref{lccform1}) can be calculated one input at a time through gradient accumulation. Moreover, for each given input, the standard backpropagation algorithm can be used to calculate the partial derivatives for all parameters of the network simultaneously. 

Closed-form expressions for the gradients $\alpha_{i,j},\,i\in\{1,\ldots,n\},\,j\in\{1,\ldots,6\}$ follow from cumbersome but basic calculus. To provide the final expressions, for any $i\in\{1,\ldots,n\}$, let
\begin{align}
\xi_i \triangleq \sum_{\substack{j=1 \\ j \neq i}}^n \Bigl[\exp(\tfrac{1}{\tau}s(u_i,u_j)) \!+\! \exp(\tfrac{1}{\tau}s(u_i,v_j)) \Bigr],
\end{align}
We also define
\begin{align}
\label{algebra1}
s'(u,v)\triangleq \frac{\partial s(u,v)}{\partial u} = \frac{s(u,v)}{\|u\|^2} u + \frac{1 }{\|u\|\|v\|} v,
\end{align}
using which we can provide the closed-form expressions
\begin{align}
 \nonumber n \tau  & \frac{\partial L(\mathbf{u},\mathbf{v};\tau)}{\partial u} = -s'(u_{\ell},v_{\ell})  + \\  \label{algebra2} & 
\sum_{\substack{i=1 \\ i \neq {\ell}}}^n \Bigl[ \Bigl(\frac{1}{ \xi_{\ell}}+\frac{1}{ \xi_{i}}\Bigr) \exp(\tfrac{1}{\tau}s(u_{\ell},u_i))s'(u_{\ell},u_i) + 
\exp(\tfrac{1}{\tau}s(u_{\ell},v_i))s'(u_{\ell},v_i)\Bigr], \\
\label{algebra3} n \tau & \frac{\partial  L(\mathbf{u},\mathbf{v};\tau) }{\partial v}  = 
-s'(v_{\ell},u_{\ell}) + \sum_{\substack{i=1 \\ i \neq \ell}}  \frac{1}{ \xi_i} \exp(\tfrac{1}{\tau}s(v_{\ell},u_{i}))s'(v_{\ell},u_{i})
\end{align}
for the gradients of $L(\mathbf{u},\mathbf{v};\tau)$. In addition, with regards to the entropies that appear in the loss function, we can evaluate, for any $\ell\in\{1,\ldots,d\}$,
\begin{align}
\label{algebra4}
    \frac{\partial H(\mathbf{u})}{\partial u_{\ell}} 
        =  \sum_{i=1}^{d} \frac{\|u_i\|_1-\mathbf{1}(i=\ell)\|\mathbf{u}\|_1}{\|\mathbf{u}\|_1^2}\left(1+\log \frac{\|u_i\|_1}{\|\mathbf{u}\|_1}\right)\mathrm{sign}(u_{\ell}).
\end{align}
As mentioned, formulae (\ref{algebra1})-(\ref{algebra4}) can be verified using straightforward calculations. Now, $\alpha_{i,1}$ and $\alpha_{i,2}$ in (\ref{alphas123}) can be calculated via (\ref{algebra2}). For $\alpha_{i,3}$, we can first use the identity
\begin{align}
\label{laksjdlaksd}
\left[ \alpha_{1,3} \cdots \alpha_{n,3} \right] \!=\! \frac{1}{2}\left[\frac{\partial L(\mathbf{c}^a, \mathbf{c}^b; \tau_C)}{\partial c_1^a} \cdots \frac{\partial L(\mathbf{c}^a, \mathbf{c}^b; \tau_C)}{\partial c_d^a} \right]^{\dagger}\!\!\!\!,\!\!\!\!
\end{align}
where $c_1^{\sigma},\ldots,c_n^{\sigma}$ represent the columns of $\mathbf{c}^a$. Now, the right hand side of (\ref{laksjdlaksd}) can be calculated through (\ref{algebra3}). Likewise, $\alpha_{i,4},\alpha_{i,5}$, and $\alpha_{i,6}$ can be evaluated via (\ref{algebra2}), (\ref{algebra3}),  (\ref{algebra4}), and the transposition method in (\ref{laksjdlaksd}).

%% file: 06-conc.tex
\section{Conclusions}
\label{secconclusions}
We have presented Federated Momentum Contrastive Clustering (FedMCC), a deep federated clustering model based on contrastive learning. 
FedMCC clusters a set of data points distributed over multiple clients and is trained by contrastive loss on a momentum network. 
Our method can be viewed as a fully unsupervised federated clustering scheme, extending existing research that primarily deals with self-supervised or semi-supervised representation learning. 
Our framework provides a distributed way to learn representations specialized for clustering. 
The extensive experiments demonstrate that the FedMCC provides good performance on the federated clustering scenario and achieves state-of-the-art clustering accuracy on several centralized datasets. In addition, FedMCC can also be adapted to linear evaluation and semi-supervised settings, achieving state-of-the-art results as well. 
Our approach opens up future interesting research directions such as clustering data with a large amount of clusters in the federated setting or improving the overall clustering accuracy. 